\pdfoutput=1

\documentclass[11pt]{article}

\usepackage[final]{acl}

\usepackage{times}
\usepackage{latexsym}

\usepackage[T1]{fontenc}

\usepackage[utf8]{inputenc}

\usepackage{microtype}

\usepackage{inconsolata}

\usepackage{graphicx}

\usepackage{multirow}
\usepackage{booktabs}
\usepackage{enumitem}

\usepackage{algorithm}
\usepackage{algorithmic}
\usepackage[utf8]{inputenc} 
\usepackage[T1]{fontenc}    
\usepackage{hyperref}       
\usepackage{url}            
\usepackage{booktabs}       
\usepackage{amsfonts}       
\usepackage{nicefrac}       
\usepackage{microtype}      
\usepackage{xcolor}         
\usepackage{enumitem}
\usepackage{listings}
\usepackage{graphicx}  
\usepackage{float}  
\usepackage{subcaption} 
\usepackage{algorithm}
\usepackage{algorithmic}
\usepackage{placeins}
\usepackage{amsmath}
\usepackage{adjustbox}
\usepackage{booktabs} 
\usepackage{url}
\usepackage{newfloat}
\usepackage{listings}

\title{QuantAgents: Towards Multi-agent Financial System via \\ Simulated Trading}

\author{
  \textbf{Xiangyu Li}\textsuperscript{1}\thanks{Equal contribution.}, 
  \textbf{Yawen Zeng}\textsuperscript{1}\footnotemark[1], 
  \textbf{Xiaofen Xing}\textsuperscript{1}, 
  \textbf{Jin Xu}\textsuperscript{1,2},
  \textbf{Xiangmin Xu}\textsuperscript{3,1}\thanks{Corresponding Author.}\\
  \textsuperscript{1}South China University of Technology \\
  \textsuperscript{2}Pazhou Lab \\
  \textsuperscript{3}Foshan University \\
  {\{65603605lxy, yawenzeng11\}}@gmail.com, xmxu@scut.edu.cn
}

\begin{document}
\maketitle

\begin{abstract}

In this paper, our objective is to develop a multi-agent financial system that incorporates \textbf{simulated trading}, a technique extensively utilized by financial professionals. While current LLM-based agent models demonstrate competitive performance, they still exhibit significant deviations from real-world fund companies. A critical distinction lies in the agents' reliance on ``post-reflection'', particularly in response to adverse outcomes, but lack a distinctly human capability: long-term prediction of future trends. Therefore, we introduce \textbf{QuantAgents}, a multi-agent system integrating simulated trading, to comprehensively evaluate various investment strategies and market scenarios without assuming actual risks. Specifically, QuantAgents comprises four agents: a simulated trading analyst, a risk control analyst, a market news analyst, and a manager, who collaborate through several meetings. Moreover, our system incentivizes agents to receive feedback on two fronts: performance in real-world markets and predictive accuracy in simulated trading. Extensive experiments demonstrate that our framework excels across all metrics, yielding an overall return of nearly 300\% over the three years (\url{https://quantagents.github.io/}).

\end{abstract}

\section{Introduction}
In the data-driven era, the ascendancy of artificial intelligence has sparked transformative changes within the financial area \cite{kou2019machine}. Advancements in large language models (LLMs), notably exemplified by systems like FinGPT \cite{yang2023fingpt} and FinReport \cite{li2024finreport}, are significantly elevating the automation and intelligence levels in financial analysis and decision-making. These frameworks not only enhance the scope and profundity of financial analysis but also facilitate the generation of comprehensive financial reports. Particularly noteworthy are LLM-based agent systems such as FinAgent \cite{zhang2024multimodal}, which possess the capacity to emulate human decision-making processes. These systems demonstrate prowess in iterative self-improvement via tools, memory, and reflection capabilities, thereby enabling them to execute intricate financial operations adeptly \cite{yang2023autogpt}.

\begin{figure}[t!]
    \centering
    \label{visual1}
    \includegraphics[width=0.87\linewidth]{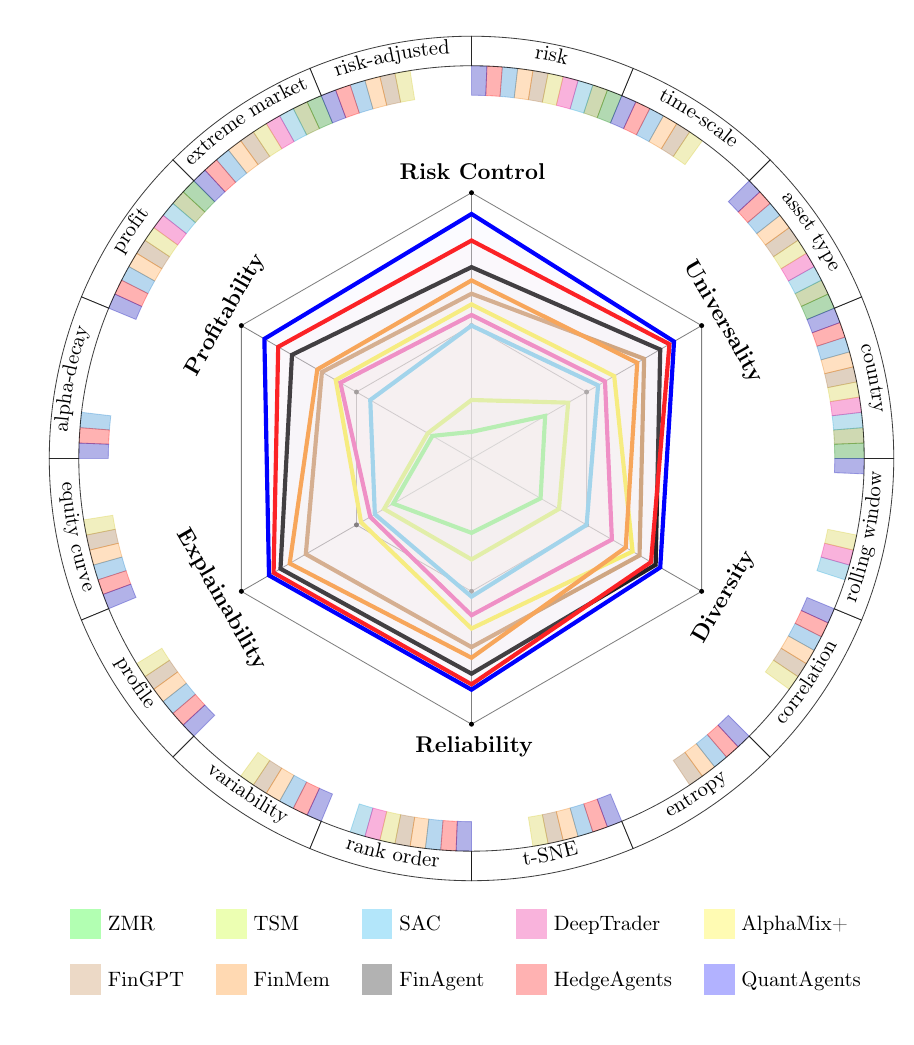}
    \vspace{-0.5cm}
    \caption{Our method has surpassed all baselines on the PRUDEX~\cite{sun2023prudexcompass} benchmark.} 
    \label{visual1}
    \vspace{-0.5cm}
\end{figure}

\begin{figure*}[t!]
    \centering
    \includegraphics[width=0.97\linewidth]{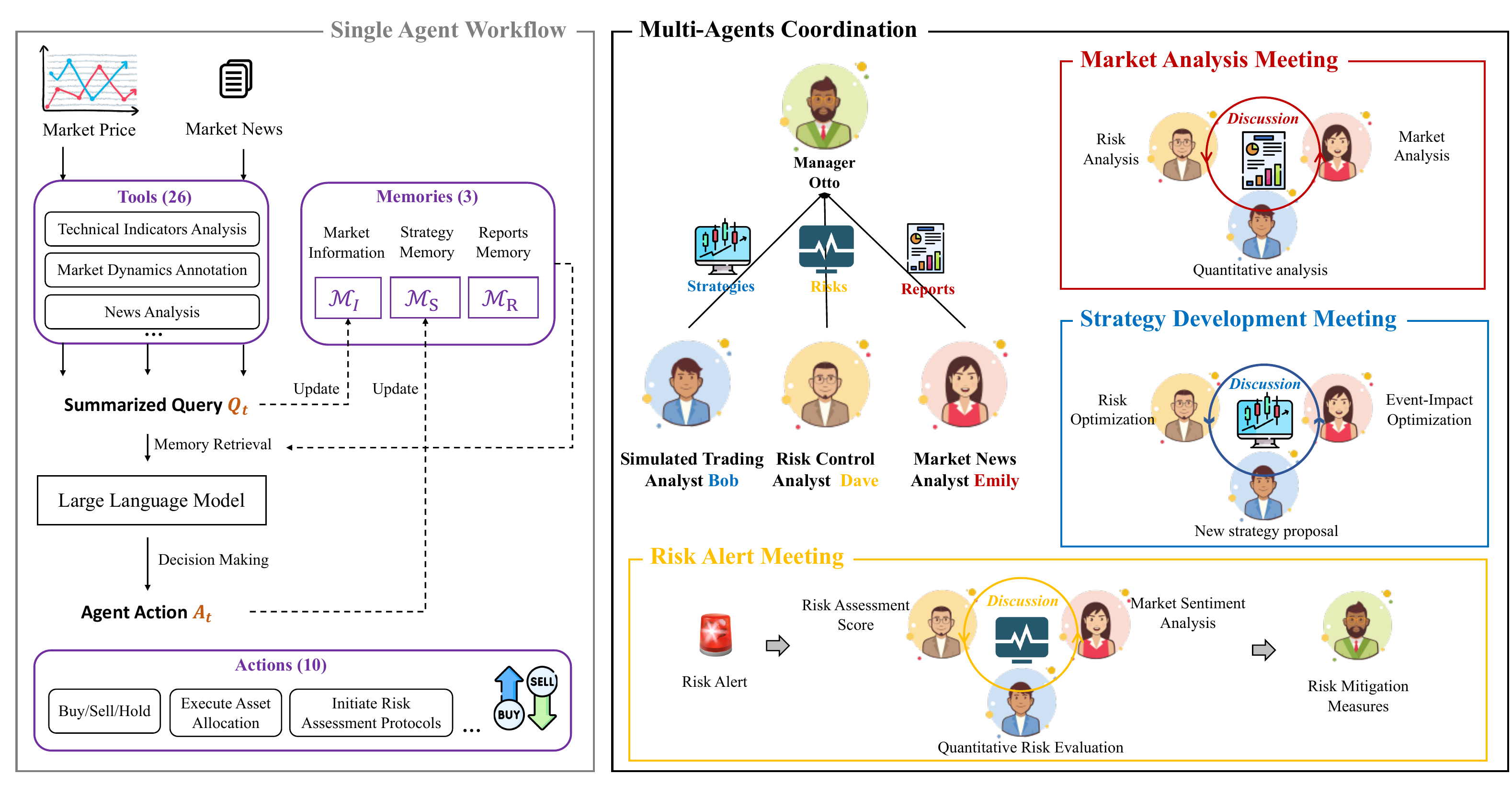}
    \vspace{-0.3cm}
    \caption{The workflow of QuantAgents, which is equipped with 26 tools, 3 types of memory to execute 10 actions. Furthermore, three meetings, i.e., {\color{red}market analysis}, {\color{blue}strategy development}, {\color{olive}risk alert} meeting will assist in decision-making (e.g., buy).}
    \label{fig:framework}
    \vspace{-0.5cm}
\end{figure*}

However, despite the strong performance of these agent-based systems in evaluations, significant disparities persist between their operational pipeline and those of real-world fund companies. A critical distinction lies in the agents' reliance on ``post-reflection'', where thinking and learning occur after events, particularly in response to adverse outcomes \cite{yang2023autogpt,park2023generative}. While this approach aids in learning from past errors, it overlooks a distinct human capability: \textit{long-term prediction of future trends}. This prediction and response to future events are pivotal in financial markets \cite{bankivestment}. Financial practitioners recognize that while market fluctuations are frequently unpredictable, sound investment decisions hinge on forward-looking market analysis. Consequently, \textbf{simulated trading} serves as an invaluable predictive instrument widely embraced by financial experts, augmenting their daily operations. This tool enables practitioners to experiment with diverse investment strategies devoid of actual risks, thereby enhancing their comprehension to market dynamics. 

Therefore, this paper introduces an innovative multi-agent financial system named \textbf{QuantAgents}, designed to achieve long-term forecasting by simulated trading. This system not only learns from actual market but also anticipates and adjusts to market fluctuations through virtual trading environments. Specifically, QuantAgents comprises four agents: a simulated trading analyst, a risk control analyst, a market news analyst, and a manager, who collaborate through several meetings. Moreover, this system incentivizes agents to receive feedback on two fronts: performance in real markets and predictive accuracy in simulated trading. This dual reward mechanism aims to encourage agents to make more precise and forward-thinking decisions in intricate and dynamic financial markets. Through this approach, we hope narrow the gap between LLM-based agents and human financial experts, offering fresh perspectives and tools for advancing the financial industry in the future.

\begin{itemize}[leftmargin=*]
\item{To the best of our knowledge, this paper represents the first endeavor in developing a multi-agent financial trading system integrated with simulated trading, configured similarly to that of human quant traders.} 
\item{We design a dual reward mechanism to coordinate agent behaviors, i.e., rewards from the real market and rewards from simulated trading. In this way, agents are encouraged to make more forward-looking decisions within the complex and dynamic financial markets.}
\item{Extensive experiments demonstrate that our framework excels across all metrics, yielding an overall return of nearly 300\%. Meanwhile, we will release all datasets and codes for the convenience of the research community\footnote{https://quantagents.github.io/}.}
\end{itemize}

\section{Related Work}
\subsection{LLM-based financial system}
Quantitative finance is an interdisciplinary field that integrates finance with mathematical and statistical methods to address complex financial challenges \cite{kou2019machine,kanamura2021pricing}. With the advent of large language models (LLMs), an increasing number of researchers are leveraging cutting-edge technologies in finance. \citet{yang2023fingpt} proposed FinGPT, which enables a thorough understanding of financial events and facilitates news analysis. \citet{li2024finreport} introduced FinRport, a framework that amalgamates diverse information to generate comprehensive financial reports on a regular basis. Compared to conventional models \cite{YU2011367,wang2021deeptrader}, these LLM-based approaches improve the accuracy and efficiency of market forecasting. However, these methods have yet to be fully learn from real-world fund companies, and essential components such as simulated trading have not been included.

\subsection{Multi-Agent Framewrok}
LLM-based agent systems, leveraging their cognitive and generative capabilities, have the ability to perform a range of complex tasks, including knowledge integration, information retention, logical reasoning, and strategic planning \cite{sumers2023cognitive,pan2023llms,chen2025combatvla,gu2025mobiler1}. Furthermore, initiatives based on multi-agent systems, such as ``The Sims'' from Stanford University \cite{park2023generative}, have demonstrated the formidable power of collective intelligence. Through the collaboration of multiple agents, multi-agent systems are expected to make significant contributions in fields such as finance \cite{zhang2024multimodal}, offering innovative approaches and sophisticated solutions for complex challenges \cite{hong2023metagpt,wu2023bloomberggpt}.

\section{Proposed Method}
\subsection{Preliminaries}
QuantAgents is a multi-agent system designed to manage a fund company operating with NASDAQ-100 index components\footnote{https://www.nasdaq.com/solutions/nasdaq-100}. 
The inputs for this system include financial data such as stock prices, financial news, and company financial reports, which guide the execution of trading actions in both simulated and real-world environments.
\subsection{Overall Framework}
QuantAgents comprises four specialized agents, each contributing to different aspects of fund management, as presented in Table \ref{tab:agents}.
These agents collaborate by participating in various meetings to assist manager Otto in decision-making. Among these meetings, market analysis meetings are held weekly to produce market reports, while strategy analysis meetings also occur weekly, focusing on enhancing investment strategies through simulated trading. Additionally, risk alert meetings are convened are triggered as needed.

\begin{table}[t]
\centering
\caption{QuantAgents comprises four agents.} 
\vspace{-0.2cm}
\resizebox{0.47\textwidth}{!}{
\begin{tabular}{l|l|l}
\hline
\textbf{Agent} & \textbf{Profession}     & \textbf{Responsibility} \\ \hline
Otto           & Manager                 & Executes Decisions      \\ \hline
{\color{blue}Bob}            & Simulated Trading Analyst        & Testing Strategies      \\ \hline
{\color{olive}Dave}           & Risk Control Analyst    & Evaluates Risks         \\ \hline
{\color{red}Emily}          & Market News Analyst & Provides Reports        \\ \hline
\end{tabular}}
\vspace{-0.2cm}
\label{tab:agents}
\end{table}

\subsection{Definitions of Single Agent}
\subsubsection{Tools.}
Each agent in QuantAgents is equipped with a set of financial analysis tools $\mathcal{T}$, comprising 26 distinct tools such as Technical Indicator Analysis, News Event Extraction, and Portfolio Stress Testing\footnote{For a detailed introduction to each component of agents, please refer to our Appendix.}.

\subsubsection{Actions.}
Each agent is defined by a detailed profile that specifies its description and permissions, ensuring a well-defined operational scope. The actions $\mathcal{A}$ include 10 types, such as Buy, Sell, and Hold. Each agent's profile specifies its description and permissions, defining its operational scope.

\subsubsection{Memories.}

The memory system $\mathcal{M}$ of each agent consists of three types: Market Information Memory ($\mathcal{M}_{I}$), which stores historical data including stock prices, financial news, and economic indicators; Strategy Memory ($\mathcal{M}_{S}$), which contains analysis of strategies in both simulated trading and real-world trading; and Report Memory ($\mathcal{M}_{R}$), which comprise in-depth analyses of markets, industries, and companies.

\subsubsection{Single Agent Workflow.}
Our QuantAgents employs a reflection-driven decision-making process, integrating LLM-based agents into a reinforcement learning framework. This workflow includes memory retrieval, decision making, and reflection update.

Firstly, we retrieve reliable experiential memories to augment decision-making. At time step $t$, a summarized query $Q_t$ is compiled from inputs (e.g., stock prices, financial news) and used to retrieve $K=10$ similar cases $\mathcal{M}_{ret}$ from the memory set $\mathcal{M} = \{\mathcal{M}_{I}, \mathcal{M}_{S}, \mathcal{M}_{R}\}$.

Based on the retrieved experiences $\mathcal{M}_{ret}$, we redefine a reinforcement learning framework to pursue the optimal investment strategy $\mu_t$.
\vspace{-0.3cm}
\begin{equation}\label{Eqn:strategy}
\begin{split}
    \pi_{\theta^*} = \arg\max_{\pi_{\theta}} \mathbb{E}_{\pi_{\theta}} [&\sum_{t=0}^T \gamma (r_t) \mid s_t = s, \mu_t = \mu ],
\end{split}
\end{equation}
where $r_t$ is a reward from real-world trading, $\gamma$ is the discount factor and the state $s_t$ includes market indicators and other relevant financial data. Actions $a_t$ are determined by:
\begin{equation}
\begin{split}
    \pi(a_t \mid s_t, \mu_t) \equiv \mathcal{D}(&LLM(\phi_D(s_t,\mathcal{M}_{ret}, \mu_t))),
\end{split}
\end{equation}
where $\mu_t$ is the trading strategy. The prompt template $\phi_D(\cdot)$ is meticulously designed, and fed into an LLM, then the response from the LLM is processed by the parsing function $\mathcal{D}(\cdot)$ to obtain an action. The policy $\pi_{\theta}$ is updated using gradient ascent. Finally, we update the $\mathcal{M}_{I}$ in real time. The all market information (e.g., stock prices), the summarized query $Q_t$, and reflections on the action are all continuously updated in $\mathcal{M}_{I}$ to guide future decisions.
A simplified prompt is as follows\footnote{The templates will change based on different tasks indicated by $\phi_{(\cdot)}$, fully disclosed in the Appendix.},
\vspace{-0.3cm}
\begin{center}
\fcolorbox{black}{gray!10}{\parbox{1\linewidth}{
<Prompt Template> \\
You are \{Dave Profile\}. The market environment today includes \{Prices\}, \{News\}. Through financial analysis tools, \{Tool Results\} can be obtained. The output format should be JSON, such as \{Examples\}.
}}
\end{center}

\subsection{Coordination of Multi-Agents}
To simulate real-world fund companies, functions such as simulated trading and market reports are integrated into our framework with collaboration among the four agents. In this section, we will provide a detailed introduction to the three types of collaborative meetings, followed by the decision-making process employed by manager Otto.

\subsubsection{Market Analysis Meeting}
The Market Analysis Meeting, scheduled weekly following the last trading day, integrates the expertise of Emily, Bob, and Dave, to generate a comprehensive market report covering news analysis, industry analysis, individual stock analysis and more. The collaborative approach ensures a balanced perspective, combining qualitative insights with quantitative rigor.

\begin{itemize}[leftmargin=*]
\item{Emily first uses the FinReport tool \cite{li2024finreport} to conduct an overall analysis based on historical data and news inputs, encompassing economic indicators, global trends, and geopolitical influences.
 } 
\item{Bob will then provide a quantitative analysis based on Emily's report, utilizing statistical tools including trend forecasting \cite{CHAUDHARI2023103293}, factor analysis \cite{FAMA20151}, and other relevant methods. The analysis will incorporate historical data and news inputs to evaluate market trends, industry performance, and individual stock metrics.
}
\item{Dave will deliver a risk analysis using the Volatility Assessment Tool \cite{mieg2022volatility}, focusing on market volatility, risks within specific industries, and vulnerabilities of individual stocks.
}
\end{itemize}

The conclusions drawn by the three analysts will be fed into the LLM in a prompt template $\phi_R(\cdot)$, resulting in a comprehensive market report, which is then stored in the Reports Memory $\mathcal{M}_{R}$ for future decision.

\begin{figure}[t!]
    \vspace{-0.3cm}
    \centering
    \includegraphics[width=0.99\linewidth]{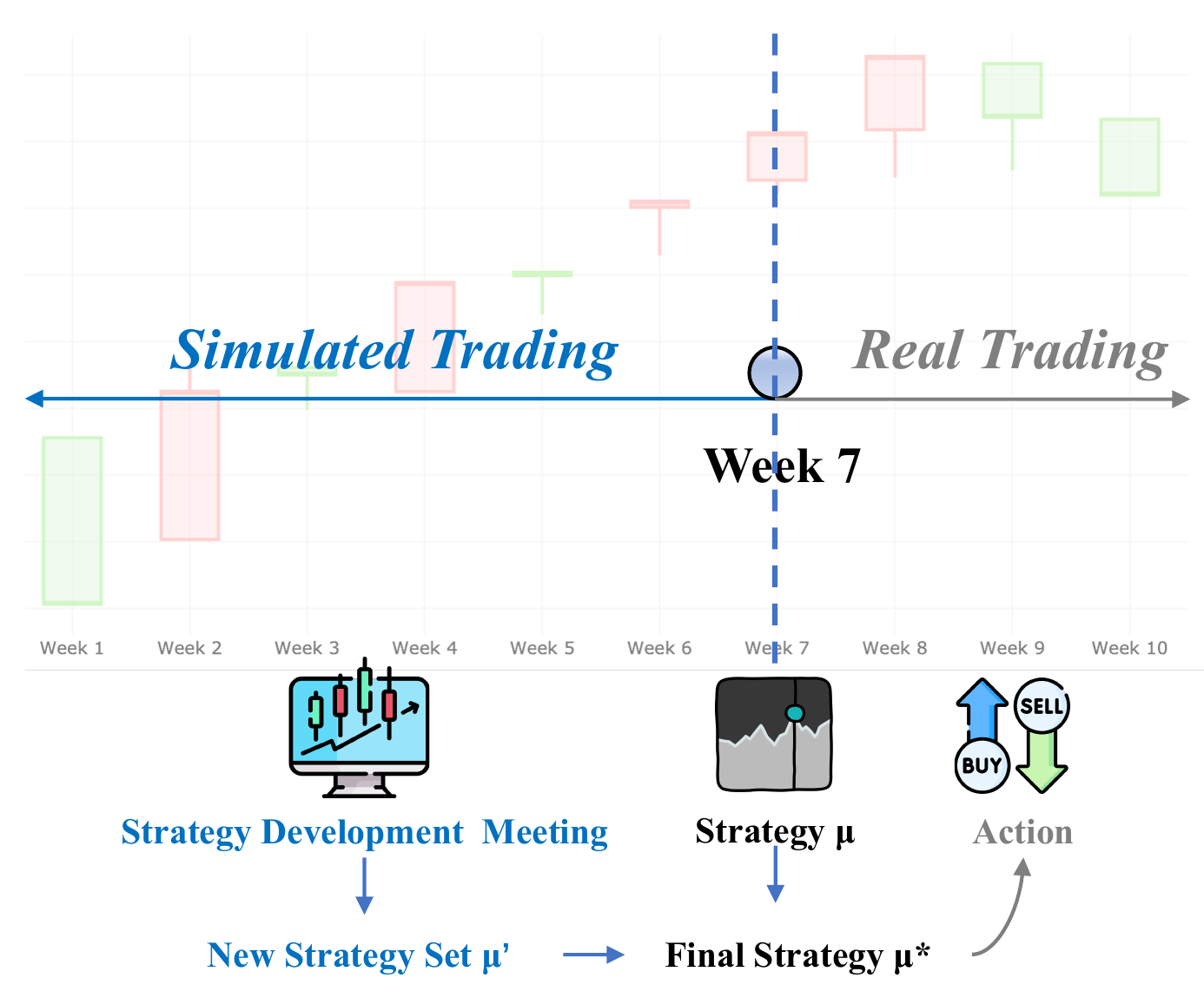}
    \caption{Our framework has been optimized to obtain rewards from both simulated and real-world trading.}
    \label{fig:simulated}
    \vspace{-0.3cm}
\end{figure}

\subsubsection{Strategy Development Meeting}
The strategy analysis meeting is held weekly after the last trading day. This meeting is crucial for implementing simulated trading to test new strategies $\mu_{t}^{'}$. During this meeting, Bob is responsible for testing new strategies, while Emily and Dave provide advice on market conditions and risk management. 
This meeting ensures that all new strategies are rigorously evaluated and refined before deployment.
\begin{itemize}[leftmargin=*]
\item{Bob will conduct simulated trading to test all potential new strategies, and outline their characteristics along with possible optimization approaches. As depicted in Figure \ref{fig:simulated}, Bob will utilize statistical tools such as the Simulation Optimization Toolkit \cite{SIMOPT2023} and the Strategy Analysis Suite \cite{STRAT2023}, along with all the data prior to the current time (e.g. Week 7) to undertake a simulated trading backtest. Bob will select the strategies that perform best in the historical data to form a new strategy set $\mu^{'}$ \footnote{We construct a strategy pool consisting of multiple index permutations, detailed in the Supplementary Material.}
}.

\item{Dave performs a risk analysis on the new strategy using the RiskAnalyzer toolkit \cite{yang2020qlibaiorientedquantitativeinvestment} , and suggests optimization directions from risk management.
 

}
\item{Emily conducts market analysis on the new strategy and proposes optimization directions based on market events. 

}
\end{itemize}

Finally, the three analysts consolidate their findings, encompassing the detailed characteristics and risk assessments of the new strategy \(\mu_{t+1}\), into the strategy memory $\mathcal{M}_{S}$ to aid action decisions.


\subsubsection{Risk Alert Meeting}
The risk alert meeting is triggered by a risk threshold to mitigate investment risk, which is defined as the combination of Portfolio Beta ($\beta_p$), Liquidity Ratio ($LR$), Sector Exposure ($SE_j$), and Volatility ($\sigma_p$) as follows,
\begin{equation}
R_{score} = w_1\beta_p + w_2(\frac{1}{LR}) + w_3\max(SE_j) + w_4\sigma_p
\end{equation}
where $w_i$ are risk factor weights. This meeting will be triggered once $R_{score} > 0.75$.

\begin{itemize}[leftmargin=*]
\item{The Risk Score Assessment tool \cite{MarketRiskAnalysis2008} is used by Dave to get a comprehensive risk analysis, including portfolio Beta value, Value at Risk (VaR), and sector concentration.}
\item{Bob conducts quantitative analysis using the StressTestPro tool \cite{KOLIAI20161}, which takes historical data and stress scenarios as inputs and outputs a risk severity score $\eta \in [0,1]$, helping to evaluate potential market impacts.

}
\item{
Emily uses the SentimentAnalyzer tool \cite{araci2019finbertfinancialsentimentanalysis} to analyze market sentiment $\tau \in [-1,1]$ for high-risk assets. This tool processes financial news and social media data to generate a sentiment score reflecting the market's positive, neutral, or negative stance.
}
\end{itemize}

Ultimately, the conclusions drawn by the three agents mentioned above are integrated by Otto to make decisions regarding high-risk situations. The modified policy $\pi_{\theta^*}^{risk}$ is:

\vspace{-0.3cm}
\begin{equation}
\begin{split}
\pi_{\theta^*}^{risk} = \arg\max_{\pi_{\theta}} \mathbb{E}_{\pi_{\theta}} [&\sum_{t=0}^T \gamma ((1-\lambda)r_t + \lambda r_t^{risk})],
\end{split}
\end{equation}
where $\lambda$ is a risk adjustment factor, and the risk-based reward is $r_t^{risk}= f(R_{score}, \eta, \tau)$.
This approach ensures that our QuantAgents remains responsive to urgent risk situations while maintaining its long-term learning capabilities.

\subsubsection{Decision Making from Otto.}
Manager Otto synthesizes all information and executes trading actions according to the optimal strategy. With the introduction of simulated trading, Otto now receives rewards from two distinct sources: real- world trading and the simulated trading of strategies. As depicted in Figure \ref{fig:simulated}, Otto conducts the action in accordance with the policy $\mu$ at the present moment (e.g., Week 7), along with the newly available policy $\mu^{'}$, as the final policy $\mu^{*}$.

Consequently, Eqn. \ref{Eqn:strategy} will be revised as follows:
\begin{align}
   \pi_{\theta^*} &= \arg\max_{\pi_{\theta}} 
   \mathbb{E}_{\pi_{\theta}} \Biggl[\sum_{t=0}^T \gamma \bigl(
   w_t^{sim} r_t^{sim} \nonumber \\
   &\quad + w_t^{real} r_t^{real}\bigr)\Biggr]
\end{align}

where $r_t^{sim}$ is the reward from the simulated trading, and $r_t^{real}$ is the reward from the real-world trading environment. $w_t^{sim}$ and $w_t^{real}$ are adaptive weights.
The adaptive weights are updated based on the relative performance:
\begin{equation}
\begin{aligned}
w_t^{sim} &= \sigma\!\left(\frac{\sum_{i=t-n}^t r_i^{sim}}
{\sum_{i=t-n}^t (r_i^{sim} + r_i^{real})}\right), \\
w_t^{real} &= 1 - w_t^{sim}
\end{aligned}
\end{equation}
where $\sigma(\cdot)$ is the sigmoid function and $n$ is the number of recent time steps considered.

\begin{table*}[t]
\setlength{\tabcolsep}{6pt} 
\renewcommand{\arraystretch}{1.0} 
\centering
\caption{Cumulative Returns Comparison of our QuantAgents and all baselines. \textbf{Bold} represents optimal performance, while \underline{underline} represents suboptimal.}
\resizebox{0.99\textwidth}{!}{
\begin{tabular}{llccccccccc}
\hline
\hline
Categories   & Models               & ARR(\%)        & TR(\%)          & SR            & CR             & SoR            & MDD(\%)        & VoL(\%)       & ENT           & ENB           \\ \hline
Market Index & NDX                  & 9.84           & 32.52           & 0.64          & 1.38           & 13.07          & 35.58          & \underline{1.52}    & —             & —             \\ \hline
             & MV                   & 11.3           & 37.87           & 0.72          & 3.27           & 22.05          & 64.15          & 5.79          & 1.01          & 1.02          \\
Classical    & ZMR                  & 4.19           & 13.1            & 0.63          & 2.52           & 18.43          & 72.89          & 5.82          & 1.43          & 1.09          \\
             & TSM                  & 5.68           & 18.02           & 0.64          & 3.11           & 17.27          & 58.36          & 5.65          & 1.03          & 1.07          \\ \hline
             & SAC                  & 22.14          & 82.23           & 0.84          & 2.99           & 23.63          & 40.13          & 2.85          & 1.49          & 1.11          \\
RL-based     & DeepTrader           & 32.06          & 130.29          & 1.27          & \underline{7.16}     & 30.31          & 29.16    & 2.81          & 1.88          & 1.19          \\
             & AlphaMix+            & 32.51          & 132.72          & 1.49          & 5.76           & 30.66          & 40.71          & 2.85          & \underline{2.76}    & 1.36          \\ \hline
             & FinGPT               & 36.71          & 155.52          & 1.66          & 6.34           & 42.31          & 37.99          & 2.83          & 1.94          & 1.21          \\
LLM-based    & FinMem               & 37.73          & 161.25          & 1.89          & 6.16           & 43.02          & 40.19          & 2.82          & 2.25          & 1.24          \\
             & FinAgent             & 45.31    & 206.83    & 2.25   & 6.98           & \underline{47.66}    & 38.48          & 2.92          & 2.71          & \underline{1.38}    \\
             & HedgeAgents          & \underline{49.25}    & \underline{230.39}    & \underline{2.41}    & 6.53     & 45.21   & \underline{23.65}    & 1.99   & 2.68   & 1.35 \\\hline
\textbf{Ours}         & \textbf{QuantAgents} & \textbf{58.68} & \textbf{299.55} & \textbf{3.11} & \textbf{11.38} & \textbf{66.94} & \textbf{16.86} & \textbf{1.43} & \textbf{2.97} & \textbf{1.49} \\ \hline
\multicolumn{2}{c}{\textbf{Improvement(\%)}} & 19.15          & 30.02           & 29.05         & 58.94          & 40.45          & 28.71        & 5.92          & 7.61          & 7.97          \\ 
\hline
\hline
\end{tabular}}
\label{tab:overall}
\vspace{-0.3cm}
\end{table*}

\section{Experiments}
\subsection{Datasets}
The constituents of the NASDAQ-100 from January 1, 2010, to December 31, 2023, will serve as the evaluation dataset. This dataset includes daily trading data for each stock, comprising open, high, low, and closing prices, as well as trading volume, along with 60 standard technical indicators for analysis. Moreover, we incorporated daily news updates, company financial reports, and macroeconomic policy information for each asset. These data were sourced from Yahoo Finance for market data and the Alpaca News API for texts. 

\subsection{Evaluation Metrics}
We compare all models in terms of 9 financial metrics following \cite{sun2023trademaster,qin2023earnhft}, which include 2 profit metrics: Total Return (TR), Annual Return Rate (ARR); 3 risk-adjusted profit metrics: Sharpe Ratio (SR), Calmar Ratio (CR), Sortino Ratio (SOR); 2 risk metrics: Maximum Drawdown (MDD), Volatility (VOl); and 2 diversity metrics: Entropy (ENT) and Effect Number of Bets (ENB). 
Higher values are preferred for all metrics except MDD and VOL, where lower values indicate better performance.

\subsection{Implementation Details}
The dataset will be temporal split, with data from January 1, 2010, to December 31, 2020, constituting the training set, and data from January 1, 2021, to December 31, 2023, serving as the test set. 
For LLM-based approaches like QuantAgents, we adopt ``GPT-4o-2024-05-13'' as the foundation model, setting the temperature to 0.7 to balance consistency and creativity. The memory module operates as a similarity-based storage and retrieval system, utilizing the text-embedding-3-large model\cite{openai2023textembedding3} for text vectorization. The retrieval process is configured to return the top 10 results for efficient information recall.


\begin{figure}[t]
  \centering
   \includegraphics[width=0.48\textwidth,keepaspectratio]{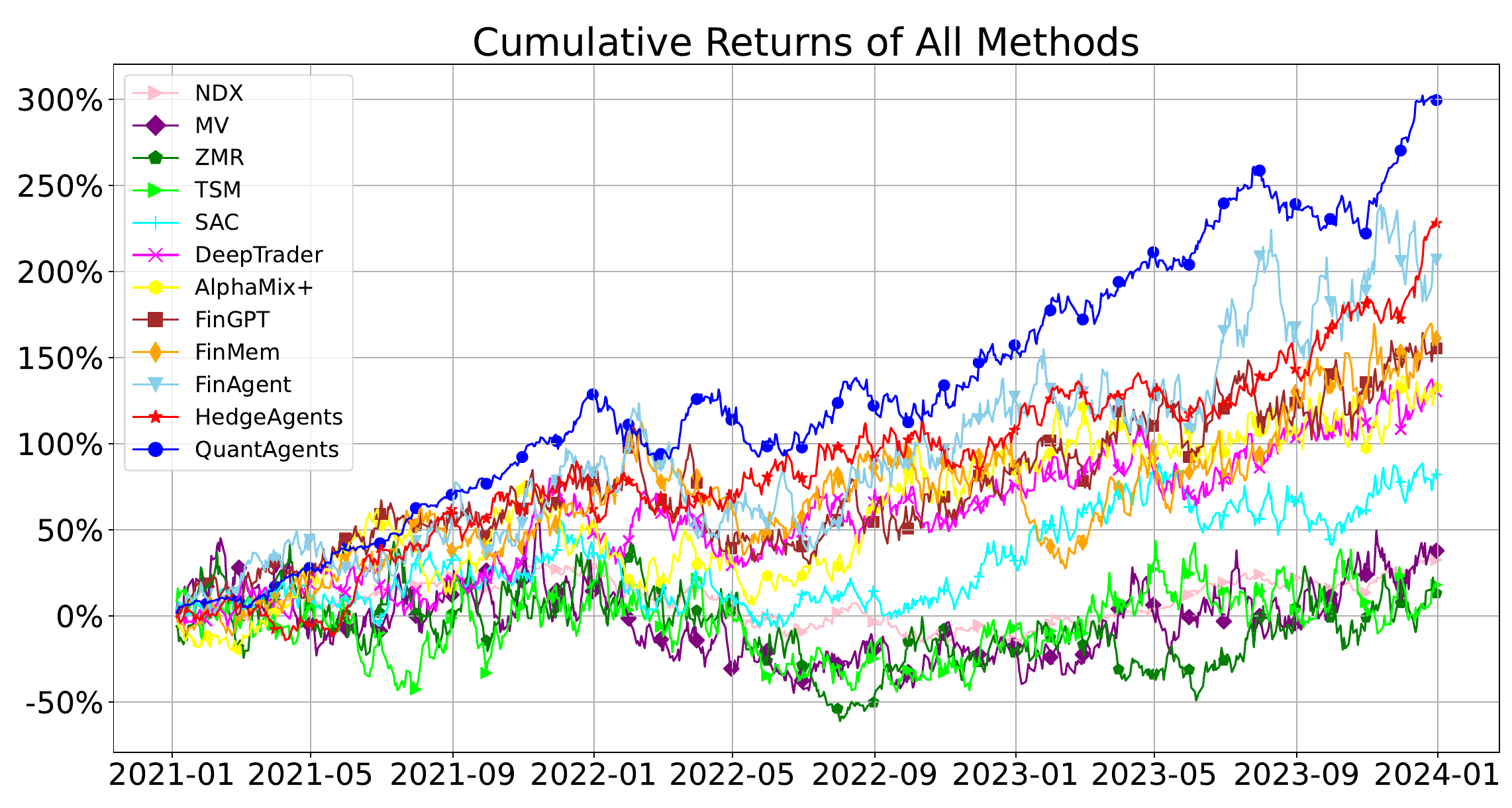}
  \caption{Cumulative Returns Comparison.
  }
  \label{fig:CRAll}
  \vspace{-0.3cm}
\end{figure}

\begin{table*}[t]
\setlength{\tabcolsep}{6pt} 
\renewcommand{\arraystretch}{1.1} 
\centering
\caption{Ablation analysis on three conference. \checkmark denote the inclusion of components.}
\vspace{-0.3cm}
\resizebox{0.99\textwidth}{!}{
\begin{tabular}{ccc|ccccccccc}
\hline
\hline
MAM & SDM & RAM & ARR(\%)        & TR(\%)          & SR            & CR             & SoR            & MDD(\%)        & VoL(\%)       & ENT           & ENB           \\ \hline
\checkmark   &     &     & 40.89          & 179.66          & 1.81          & 5.27           & 51.23          & 28.61          & 1.66          & 1.73          & 1.22          \\
    & \checkmark   &     & 43.25          & 193.97          & 1.93          & 6.27           & 53.41          & 24.99          & 1.63          & 1.89          & 1.33          \\
    &     & \checkmark   & 35.53          & 148.93          & 1.88          & 5.94           & 52.01          & 21.73          & 1.34          & 1.65          & 1.19          \\ \hline
    & \checkmark   & \checkmark   & 48.59          & 228.07          & 2.79          & 8.54           & 58.85          & 19.21          & 1.38          & 2.37          & 1.39          \\
\checkmark   &     & \checkmark   & 46.42          & 213.94          & 2.51          & 7.82           & 55.21          & 20.52          & 1.28          & 2.18          & 1.35          \\
\checkmark   & \checkmark   &     & 52.71          & 256.12          & 2.86          & 9.14           & 61.46          & 21.85          & 1.33          & 2.55          & 1.43          \\ \hline
\checkmark   & \checkmark   & \checkmark   & \textbf{58.68} & \textbf{299.55} & \textbf{3.11} & \textbf{11.38} & \textbf{66.94} & \textbf{16.86} & \textbf{1.23} & \textbf{2.97} & \textbf{1.49} \\ \hline
\hline
\end{tabular}}
    \label{tab:ablationmeeting}
    \vspace{-0.3cm}
\end{table*}

\begin{figure*}[htbp]
  \centering
   \includegraphics[width=1\textwidth,keepaspectratio]{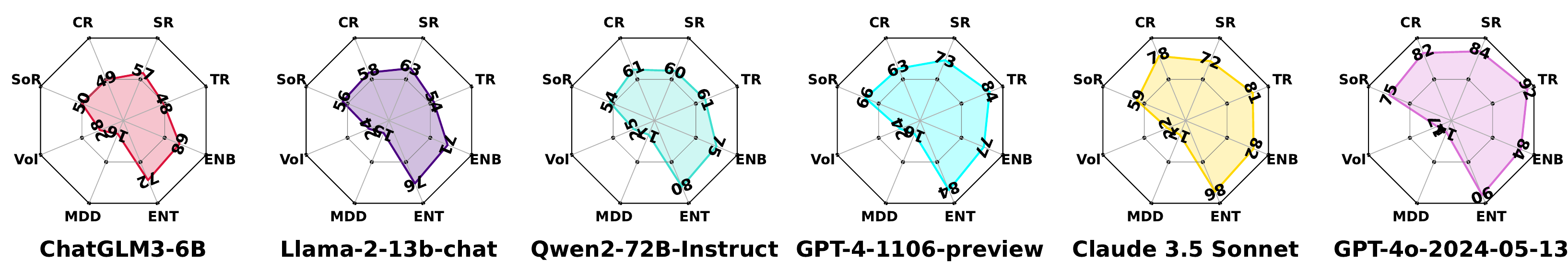}
  \vspace{-0.8cm}
  \caption{Ablation analysis on several LLM backbones, from open-source to closed-source models. The numbers presented in the figure have been normalized and converted into percentage values.}
  \label{fig:rq3}
  \vspace{-0.3cm}
\end{figure*}

\subsection{Overall Performance Comparison} 
Table \ref{tab:overall} presents a comprehensive comparison of QuantAgents against a diverse set of baseline models. These baselines include 1) three classical rule-based quantitative investment strategies: MV\cite{YU2011367}, ZMR\cite{eeckhoudt2018dual}, and TSM\cite{moskowitz2012time}; 2) three reinforcement learning-based financial agents: SAC \cite{haarnoja2018soft}, DeepTrader\cite{Wang_Huang_Tu_Zhang_Xu_2021}, and AlphaMix+\cite{sun2023prudexcompass}; 3) and four LLM-based methods: FinGPT\cite{yang2023fingpt}, FinMem\cite{yu2023finmem}, FinAgent\cite{zhang2024multimodal} and HedgeAgents\cite{li2025hedgeagentsbalancedawaremultiagentfinancial}. 
The following observations can be made: 

1) RL-based methods outperform rule-based strategies in both profitability and risk-adjusted performance. 
For instance, DeepTrader achieves an annualized return rate (ARR) of 32.06\% and a Sharpe ratio (SR) of 1.27, surpassing the best-performing classical strategy (MV) with an ARR of 11.3\% and SR of 0.72. 

2) LLM-based methods further improve upon RL-based approaches, showcasing the power of large-scale language models in financial decision-making. 
FinAgent achieves an impressive ARR of 45.31\% and SR of 2.25, significantly outperforming the top RL-based model, AlphaMix+ (ARR: 32.51\%, SR: 1.49). The cumulative returns graph Figure \ref{fig:CRAll}  clearly illustrates the superior performance trajectory of LLM-based methods, particularly from mid-2022 onwards.

3) QuantAgents, our proposed multi-agent system, demonstrates superior performance across all evaluation metrics, achieving the highest ARR (58.68\%) and SR (3.11), surpassing the best baseline (HedgeAgents) by 19.15\% and 30.02\%. 
QuantAgents also excels in risk management with the lowest MDD (16.86\%) and VoL (1.43\%), while achieving the highest portfolio diversity with ENT (2.97) and ENB (1.49), indicating a more balanced and robust investment strategy.
Figure \ref{fig:CRAll} vividly showcases QuantAgents' outstanding performance, with its cumulative returns curve consistently above all other methods, reaching approximately 300\% by the end of the test period. This remarkable performance can be attributed to the synergistic collaboration of specialized agents within QuantAgents, each contributing unique expertise in investment management, strategy development, risk control, and market analysis. 

\subsection{Ablation Study}
\subsubsection{Effectiveness of Each Meeting}

We conducted an ablation study to evaluate the contribution of each meeting module in QuantAgents. Table \ref{tab:ablationmeeting} presents the performance metrics for different combinations of Market Analysis Meeting (MAM), Strategy Development Meeting (SDM), and Risk Assessment Meeting (RAM). We have the following observations: 1) MAM significantly enhances profitability and portfolio diversity, as evidenced by its high ARR (40.89\%) and ENT (1.73) when used alone. 
Its exclusion from two-meeting combinations reduces ARR and ENT, highlighting its critical role in trend identification and diversification. 2) SDM enhances risk-adjusted returns and portfolio efficiency, with the highest single-meeting SR (1.93) and CR (6.27). Its inclusion in two-meeting setups consistently improves these metrics. The SDM-MAM combination achieves the highest two-meeting SR (2.86) and CR (9.14), demonstrating SDM's effectiveness in strategy formulation.
3) RAM demonstrates strength in risk management and volatility reduction. Despite having the lowest standalone ARR (35.53\%), it achieves the lowest single-meeting MDD (21.73\%) and VoL (1.34\%). In two-meeting configurations,RAM improves risk metrics, 
notably reducing MDD when combined with SDM. 
4) The synergistic effect of all three meetings is evident in our QuantAgents, which outperforms all partial combinations across all metrics. It achieves the highest ARR (58.68\%), SR (3.11), and ENT (2.97), while maintaining the lowest MDD (16.86\%) and VoL (1.23\%). 
\begin{figure*}[t!]
    \centering
    \includegraphics[width=0.99\linewidth]{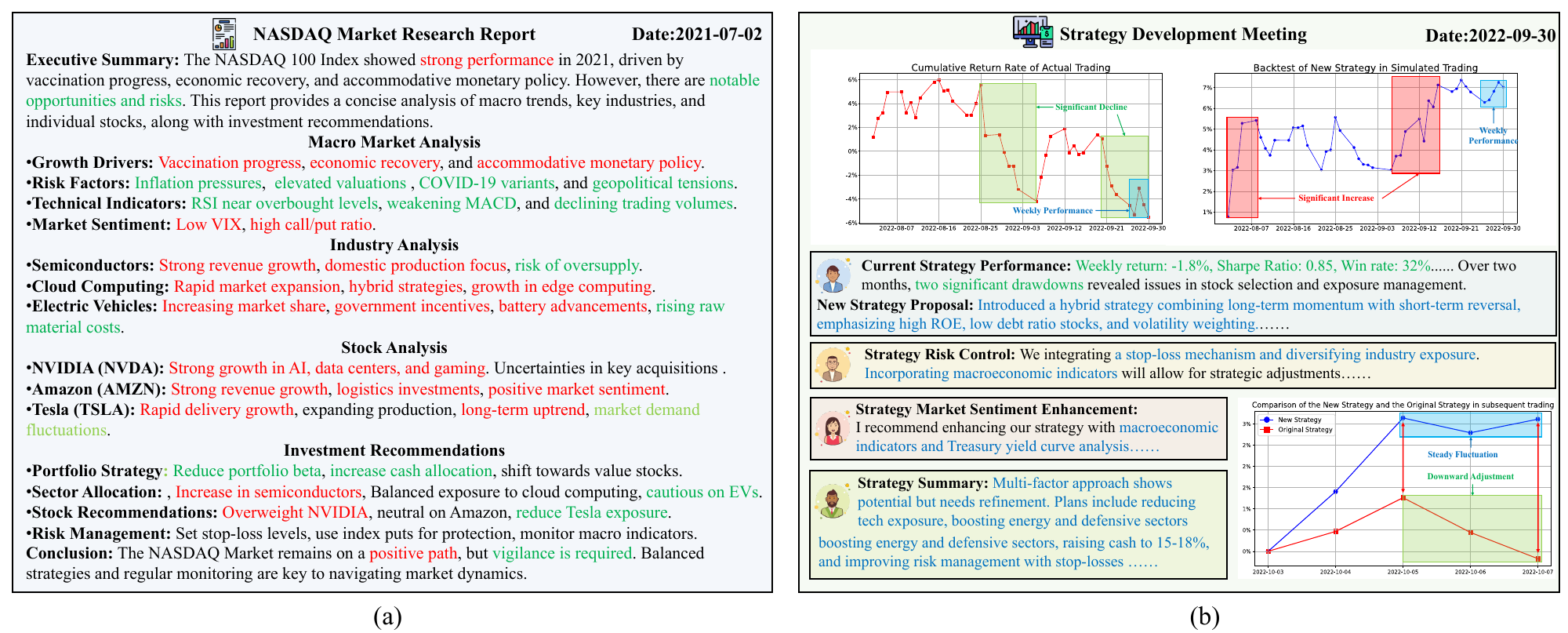}
    \vspace{-0.3cm}
    \caption{Visualizations of Market Research Report and Strategy Development Meeting.}
    \label{fig:vis}
\end{figure*}

\subsubsection{Effectiveness of LLM Backbone}
To evaluate the performance of different LLMs as the backbone, we selected 6 representative models, including ChatGLM3-6B\cite{glm2024chatglm}, Llama-2-13b-chat\cite{touvron2023llama2openfoundation}, Qwen2-72B-Instruct\cite{qwen2}, GPT-4-1106-preview\cite{openai2024gpt4},  Claude 3.5 Sonnet\cite{claude_3.5_sonnet}, and GPT-4o-2024-05-13\cite{wu2024gpt4ovisualperceptionperformance}. Each of these models serves as the brain of QuantAgents, as shown in Figure \ref{fig:rq3}. We have the following observations: 
1) QuantAgents achieves consistent performance across diverse LLM backbones, showcasing its adaptability through a robust multi-agent architecture.
2) Larger models, such as Qwen2-72B-Instruct, outperform smaller ones like Llama-2-13b-chat, likely due to their superior capacity to handle complex financial data and detect subtle market patterns.
3) Closed-source models like Claude 3.5 Sonnet outperform open-source ones, such as Qwen2-72B-Instruct, likely due to proprietary training data and advanced fine-tuning techniques enhancing their understanding of financial contexts.
Therefore, we select GPT-4o as the core of QuantAgents. Notably, our system has accumulated a total cost of \$180 over the three years, averaging only \$0.17 per day!

\subsection{Visualization}
\subsubsection{Market Analysis Meeting}
Figure \ref{fig:vis} (a) showcases the visualization of market research report dated 2021-07-02, generated after the market analysis meeting. The report encapsulates key market insights in a structured layout, progressing from an executive summary highlighting the NASDAQ 100 Index's strong 2021 performance to specific investment recommendations.

\subsubsection{Strategy Development Meeting}
Figure \ref{fig:vis} (b) presents the visualization of a strategy development meeting held on 2022-09-30. The upper left shows the current strategy's performance, marked by a -1.8\% weekly return and significant drawdowns due to issues in stock selection and exposure management. 
The upper right shows a six-month backtest of the new strategy, yielding 35.3\% return and a 1.85 Sharpe Ratio. Dave suggested mitigating risks through stop-loss mechanisms and diversified industry exposure. Emily recommended incorporating macroeconomic indicators and adjusting positions in overvalued tech stocks. Otto summarized the strategy, emphasizing risk management and dynamic adjustments. 


\subsection{Real-World Investment Performance}
We evaluated its live trading performance in the A-stock and HK-stock markets from Q3 2024 to Q1 2025. Figure~\ref{fig:live_trading} shows cumulative returns, QuantAgents delivered superior returns of 111.87\% (Sharpe Ratio: 2.02, Win Rate: 61.23\%) in A-stocks and 97.69\% (Sharpe Ratio: 1.76, Win Rate: 59.71\%) in HK-stocks, highlighting exceptional profitability and risk management across diverse market conditions.

\begin{figure}[t]
    \centering
    \includegraphics[width=0.45\textwidth]
    {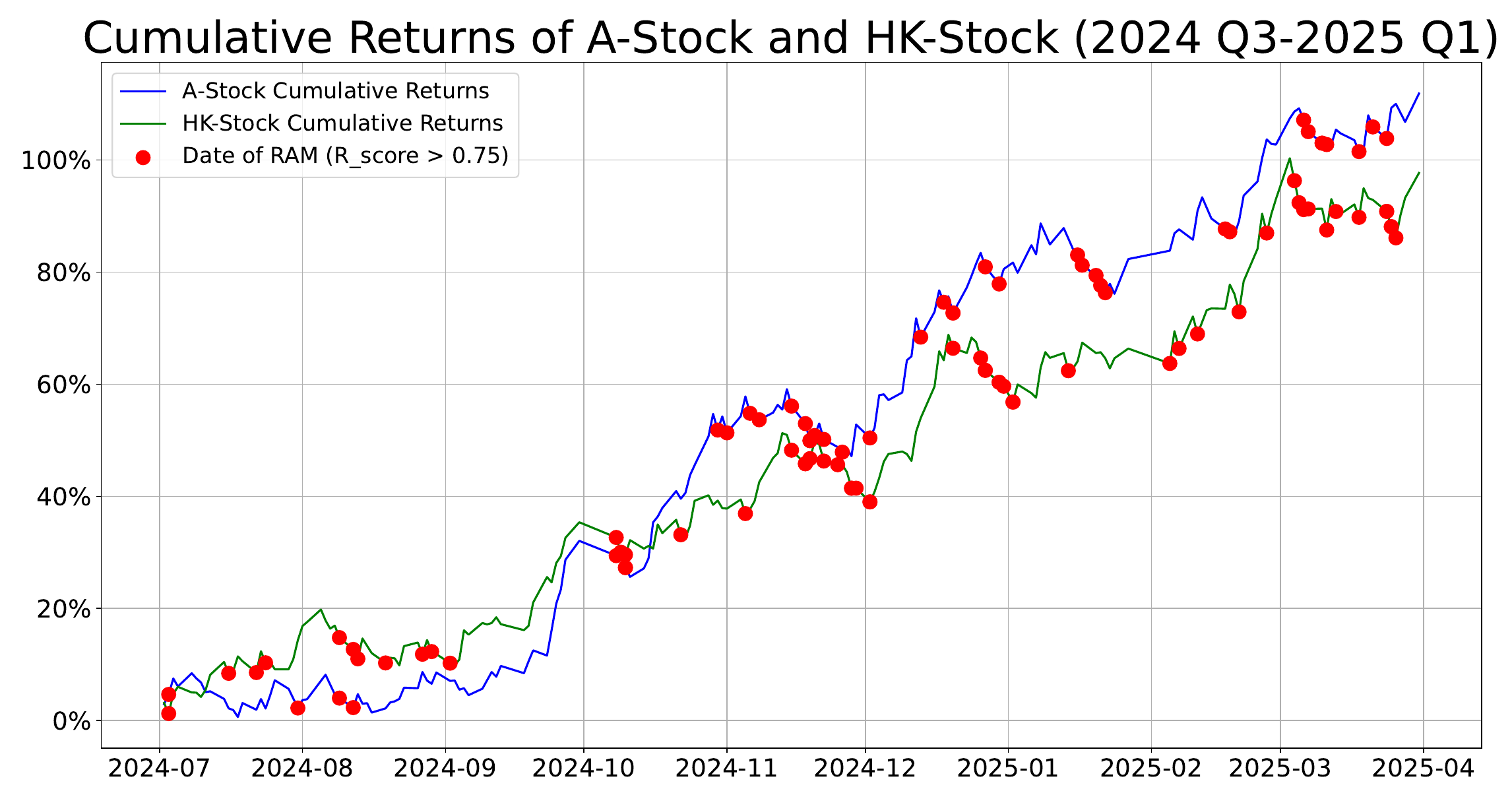}
    \vspace{-0.3cm}
    \caption{Cumulative Returns of QuantAgents during live
trading (24Q3-25Q1). RAM were held 36 times in the
A-stock and 46 times in the HK stock market.}
    \label{fig:live_trading}
\end{figure}
\section{Conclusions}
In this paper, we present a sophisticated multi-agent financial system, QuantAgents that incorporates simulated trading, configured similarly to that of human quant traders. 
Furthermore, our system encourages agents to receive feedback on their performance in the real market and their predictive accuracy in simulated trading. Compared to baselines and variations in meeting structures, 
our framework demonstrated strong performance across all metrics, resulting in an impressive overall return of nearly 300\%. 



\section*{Limitations}

We still have the following limitations: 1) In terms of generalization, our QuantAgents is modular and flexible,  designed to adapt to various market scenarios. We will explore generalization further in future work.
 2) Although backtesting may be affected by potential LLM information leakage, QuantAgents’ effectiveness is proven through impeccable live trading performance in A-stock and HK-stock markets over three quarters. We plan to validate it across global markets and report performance periodically.


\section*{Ethical Impact}
We respect intellectual property rights and comply with relevant laws and regulations. The documents in our dataset are publicly available, and we have taken careful measures to ensure that the documents in our dataset do not contain any personal sensitive information. In addition, our work is only for research purposes, not for commercial purposes.

\section*{Acknowledgments}
This work is supported in part by the National Natural Science Foundation of China (62372187), in part by the National Key Research and Development Program of China (2022YFC3601005) and in part by the Guangdong Provincial Key Laboratory of Human Digital Twin (2022B1212010004).
The authors are highly grateful to the anonymous reviewers for their careful reading and insightful comments.

\bibliography{quant}

\begin{thebibliography}{41}
\providecommand{\natexlab}[1]{#1}

\bibitem[{Alexander(2008)}]{MarketRiskAnalysis2008}
Carol Alexander. 2008.
\newblock \href {https://www.wiley.com/en-dk/Market+Risk+Analysis%2C+Volume+III%2C+Pricing%2C+Hedging+and+Trading+Financial+Instruments-p-9780470997895} {\emph{Market Risk Analysis, Volume III: Pricing, Hedging, and Trading Financial Instruments}}.
\newblock Wiley.

\bibitem[{Anthropic(2024)}]{claude_3.5_sonnet}
Anthropic. 2024.
\newblock Claude 3.5 sonnet.
\newblock Available on \url{https://claude.ai/}.
\newblock Accessed: 2024-07-15.

\bibitem[{Araci(2019)}]{araci2019finbertfinancialsentimentanalysis}
Dogu Araci. 2019.
\newblock \href {https://arxiv.org/abs/1908.10063} {Finbert: Financial sentiment analysis with pre-trained language models}.
\newblock \emph{Preprint}, arXiv:1908.10063.

\bibitem[{Buz and de~Melo(2023)}]{bankivestment}
Tolga Buz and Gerard de~Melo. 2023.
\newblock \href {https://doi.org/10.48550/ARXIV.2301.00170} {Democratization of retail trading: Can reddit's wallstreetbets outperform investment bank analysts?}
\newblock \emph{arXiv preprint}.

\bibitem[{Chaudhari and Thakkar(2023)}]{CHAUDHARI2023103293}
Kinjal Chaudhari and Ankit Thakkar. 2023.
\newblock \href {https://doi.org/10.1016/j.ipm.2023.103293} {Data fusion with factored quantization for stock trend prediction using neural networks}.
\newblock \emph{Information Processing \& Management}, 60(3):103293.

\bibitem[{Chen et~al.(2025)Chen, Bu, Wang, Wang, Wang, Guo, Zhao, Zhu, Song, Yang, Wang, and Zheng}]{chen2025combatvla}
Peng Chen, Pi~Bu, Yingyao Wang, Xinyi Wang, Ziming Wang, Jie Guo, Yingxiu Zhao, Qi~Zhu, Jun Song, Siran Yang, Jiamang Wang, and Bo~Zheng. 2025.
\newblock \href {https://arxiv.org/abs/2503.09527} {Combatvla: An efficient vision-language-action model for combat tasks in 3d action role-playing games}.
\newblock \emph{Preprint}, arXiv:2503.09527.

\bibitem[{Eeckhoudt and Laeven(2018)}]{eeckhoudt2018dual}
Louis~R. Eeckhoudt and Roger J.~A. Laeven. 2018.
\newblock \href {https://arxiv.org/abs/1612.03347} {Dual moments and risk attitudes}.
\newblock \emph{Preprint}, arXiv:1612.03347.

\bibitem[{Fama and French(2015)}]{FAMA20151}
Eugene~F. Fama and Kenneth~R. French. 2015.
\newblock \href {https://doi.org/10.1016/j.jfineco.2014.10.010} {A five-factor asset pricing model}.
\newblock \emph{Journal of Financial Economics}, 116(1):1--22.

\bibitem[{Gaikwad et~al.(2012)Gaikwad, Doan, Bossy, Baude, and Abergel}]{STRAT2023}
Abhijeet Gaikwad, Viet{\_}Dung Doan, Mireille Bossy, Fran{\c{c}}oise Baude, and Fr{\'e}d{\'e}ric Abergel. 2012.
\newblock Superquant financial benchmark suite for performance analysis of grid middlewares.
\newblock In \emph{Modeling, Simulation and Optimization of Complex Processes}, pages 103--113, Berlin, Heidelberg. Springer Berlin Heidelberg.

\bibitem[{GLM et~al.(2024)GLM, Zeng et~al.}]{glm2024chatglm}
Team GLM, Aohan Zeng, et~al. 2024.
\newblock \href {https://arxiv.org/abs/2406.12793} {Chatglm: A family of large language models from glm-130b to glm-4 all tools}.
\newblock \emph{Preprint}, arXiv:2406.12793.

\bibitem[{Gu et~al.(2025)Gu, Ai, Wang, Bu, Xing, Zhu, Jiang, Wang, Zhao, Zhang, Song, Jiang, and Zheng}]{gu2025mobiler1}
Jihao Gu, Qihang Ai, Yingyao Wang, Pi~Bu, Jingxuan Xing, Zekun Zhu, Wei Jiang, Ziming Wang, Yingxiu Zhao, Ming-Liang Zhang, Jun Song, Yuning Jiang, and Bo~Zheng. 2025.
\newblock \href {https://arxiv.org/abs/2506.20332} {Mobile-r1: Towards interactive reinforcement learning for vlm-based mobile agent via task-level rewards}.
\newblock \emph{Preprint}, arXiv:2506.20332.

\bibitem[{Ha and Mueller(2024)}]{SIMOPT2023}
Yunsoo Ha and Juliane Mueller. 2024.
\newblock \href {https://arxiv.org/abs/2408.04625} {Adaptive sampling bi-fidelity stochastic trust region method for derivative-free stochastic optimization}.
\newblock \emph{Preprint}, arXiv:2408.04625.

\bibitem[{Haarnoja et~al.(2018)Haarnoja, Zhou, Hartikainen, Tucker, Ha, Tan, Kumar, Zhu, Gupta, Abbeel et~al.}]{haarnoja2018soft}
Tuomas Haarnoja, Aurick Zhou, Kristian Hartikainen, George Tucker, Sehoon Ha, Jie Tan, Vikash Kumar, Henry Zhu, Abhishek Gupta, Pieter Abbeel, et~al. 2018.
\newblock Soft actor-critic algorithms and applications.
\newblock \emph{arXiv preprint arXiv:1812.05905}.

\bibitem[{Hong et~al.(2023)Hong, Zhuge, Chen, Zheng, Cheng, Zhang, Wang, Wang, Yau, Lin, Zhou, Ran, Xiao, Wu, and Schmidhuber}]{hong2023metagpt}
Sirui Hong, Mingchen Zhuge, Jonathan Chen, Xiawu Zheng, Yuheng Cheng, Ceyao Zhang, Jinlin Wang, Zili Wang, Steven Ka~Shing Yau, Zijuan Lin, Liyang Zhou, Chenyu Ran, Lingfeng Xiao, Chenglin Wu, and Jürgen Schmidhuber. 2023.
\newblock \href {https://arxiv.org/abs/2308.00352} {Metagpt: Meta programming for a multi-agent collaborative framework}.
\newblock \emph{Preprint}, arXiv:2308.00352.

\bibitem[{Kanamura et~al.(2021)Kanamura, Homann, and Prokopczuk}]{kanamura2021pricing}
Takashi Kanamura, Lasse Homann, and Marcel Prokopczuk. 2021.
\newblock Pricing analysis of wind power derivatives for renewable energy risk management.
\newblock \emph{Applied Energy}, 304:117827.

\bibitem[{Koliai(2016)}]{KOLIAI20161}
Lyes Koliai. 2016.
\newblock \href {https://doi.org/10.1016/j.jbankfin.2016.02.004} {Extreme risk modeling: An evt–pair-copulas approach for financial stress tests}.
\newblock \emph{Journal of Banking Finance}, 70:1--22.

\bibitem[{Kou et~al.(2019)Kou, Chao, Peng, Alsaadi, Herrera~Viedma et~al.}]{kou2019machine}
Gang Kou, Xiangrui Chao, Yi~Peng, Fawaz~E Alsaadi, Enrique Herrera~Viedma, et~al. 2019.
\newblock Machine learning methods for systemic risk analysis in financial sectors.

\bibitem[{Li et~al.(2024)Li, Shen, Zeng, Xing, and Xu}]{li2024finreport}
Xiangyu Li, Xinjie Shen, Yawen Zeng, Xiaofen Xing, and Jin Xu. 2024.
\newblock \href {https://arxiv.org/abs/2403.02647} {Finreport: Explainable stock earnings forecasting via news factor analyzing model}.
\newblock \emph{Preprint}, arXiv:2403.02647.

\bibitem[{Li et~al.(2025)Li, Zeng, Xing, Xu, and Xu}]{li2025hedgeagentsbalancedawaremultiagentfinancial}
Xiangyu Li, Yawen Zeng, Xiaofen Xing, Jin Xu, and Xiangmin Xu. 2025.
\newblock \href {https://arxiv.org/abs/2502.13165} {Hedgeagents: A balanced-aware multi-agent financial trading system}.
\newblock \emph{Preprint}, arXiv:2502.13165.

\bibitem[{Mieg(2022)}]{mieg2022volatility}
Harald~A Mieg. 2022.
\newblock Volatility as a transmitter of systemic risk: Is there a structural risk in finance?
\newblock \emph{Risk Analysis}, 42(9):1952--1964.

\bibitem[{Moskowitz et~al.(2012)Moskowitz, Ooi, and Pedersen}]{moskowitz2012time}
Tobias~J Moskowitz, Yao~Hua Ooi, and Lasse~Heje Pedersen. 2012.
\newblock Time series momentum.
\newblock \emph{Journal of financial economics}, 104(2):228--250.

\bibitem[{OpenAI(2023)}]{openai2023textembedding3}
OpenAI. 2023.
\newblock text-embedding-3-large.
\newblock Available at: \url{https://openai.com/index/new-embedding-models-and-api-updates/}.

\bibitem[{OpenAI et~al.(2024)OpenAI, Achiam et~al.}]{openai2024gpt4}
OpenAI, Josh Achiam, et~al. 2024.
\newblock \href {https://arxiv.org/abs/2303.08774} {Gpt-4 technical report}.
\newblock \emph{Preprint}, arXiv:2303.08774.

\bibitem[{Pan and Zeng(2023)}]{pan2023llms}
Keyu Pan and Yawen Zeng. 2023.
\newblock \href {https://arxiv.org/abs/2307.16180} {Do llms possess a personality? making the mbti test an amazing evaluation for large language models}.
\newblock \emph{Preprint}, arXiv:2307.16180.

\bibitem[{Park et~al.(2023)Park, O'Brien, Cai, Morris, Liang, and Bernstein}]{park2023generative}
Joon~Sung Park, Joseph~C. O'Brien, Carrie~J. Cai, Meredith~Ringel Morris, Percy Liang, and Michael~S. Bernstein. 2023.
\newblock \href {https://arxiv.org/abs/2304.03442} {Generative agents: Interactive simulacra of human behavior}.
\newblock \emph{Preprint}, arXiv:2304.03442.

\bibitem[{Qin et~al.(2023)Qin, Sun, Zhang, Xia, Wang, and An}]{qin2023earnhft}
Molei Qin, Shuo Sun, Wentao Zhang, Haochong Xia, Xinrun Wang, and Bo~An. 2023.
\newblock Earnhft: Efficient hierarchical reinforcement learning for high frequency trading.
\newblock \emph{arXiv preprint arXiv:2309.12891}.

\bibitem[{Sumers et~al.(2023)Sumers, Yao, Narasimhan, and Griffiths}]{sumers2023cognitive}
Theodore~R Sumers, Shunyu Yao, Karthik Narasimhan, and Thomas~L Griffiths. 2023.
\newblock Cognitive architectures for language agents.
\newblock \emph{arXiv preprint arXiv:2309.02427}.

\bibitem[{Sun et~al.(2023{\natexlab{a}})Sun, Qin, Wang, and An}]{sun2023prudexcompass}
Shuo Sun, Molei Qin, Xinrun Wang, and Bo~An. 2023{\natexlab{a}}.
\newblock \href {https://arxiv.org/abs/2302.00586} {Prudex-compass: Towards systematic evaluation of reinforcement learning in financial markets}.
\newblock \emph{Preprint}, arXiv:2302.00586.

\bibitem[{Sun et~al.(2023{\natexlab{b}})Sun, Qin, wentao zhang, Xia, Zong, Ying, Xie, Zhao, Wang, and An}]{sun2023trademaster}
Shuo Sun, Molei Qin, wentao zhang, Haochong Xia, Chuqiao Zong, Jie Ying, Yonggang Xie, Lingxuan Zhao, Xinrun Wang, and Bo~An. 2023{\natexlab{b}}.
\newblock Trademaster: A holistic quantitative trading platform empowered by reinforcement learning.
\newblock In \emph{Thirty-seventh Conference on Neural Information Processing Systems Datasets and Benchmarks Track}.

\bibitem[{Touvron et~al.(2023)Touvron, Martin, Stone, Albert et~al.}]{touvron2023llama2openfoundation}
Hugo Touvron, Louis Martin, Kevin Stone, Peter Albert, et~al. 2023.
\newblock \href {https://arxiv.org/abs/2307.09288} {Llama 2: Open foundation and fine-tuned chat models}.
\newblock \emph{Preprint}, arXiv:2307.09288.

\bibitem[{Wang et~al.(2021{\natexlab{a}})Wang, Huang, Tu, Zhang, and Xu}]{wang2021deeptrader}
Zhicheng Wang, Biwei Huang, Shikui Tu, Kun Zhang, and Lei Xu. 2021{\natexlab{a}}.
\newblock Deeptrader: a deep reinforcement learning approach for risk-return balanced portfolio management with market conditions embedding.
\newblock In \emph{Proceedings of the AAAI Conference on Artificial Intelligence}, volume~35, pages 643--650.

\bibitem[{Wang et~al.(2021{\natexlab{b}})Wang, Huang, Tu, Zhang, and Xu}]{Wang_Huang_Tu_Zhang_Xu_2021}
Zhicheng Wang, Biwei Huang, Shikui Tu, Kun Zhang, and Lei Xu. 2021{\natexlab{b}}.
\newblock \href {https://doi.org/10.1609/aaai.v35i1.16144} {Deeptrader: A deep reinforcement learning approach for risk-return balanced portfolio management with market conditions embedding}.
\newblock \emph{Proceedings of the AAAI Conference on Artificial Intelligence}, 35(1):643--650.

\bibitem[{Wu et~al.(2023)Wu, Irsoy, Lu, Dabravolski, Dredze, Gehrmann, Kambadur, Rosenberg, and Mann}]{wu2023bloomberggpt}
Shijie Wu, Ozan Irsoy, Steven Lu, Vadim Dabravolski, Mark Dredze, Sebastian Gehrmann, Prabhanjan Kambadur, David Rosenberg, and Gideon Mann. 2023.
\newblock Bloomberggpt: A large language model for finance.
\newblock \emph{arXiv preprint arXiv:2303.17564}.

\bibitem[{Wu et~al.(2024)Wu, Hu, Fu, Zhou, and Li}]{wu2024gpt4ovisualperceptionperformance}
Yiqi Wu, Xiaodan Hu, Ziming Fu, Siling Zhou, and Jiangong Li. 2024.
\newblock \href {https://arxiv.org/abs/2406.09781} {Gpt-4o: Visual perception performance of multimodal large language models in piglet activity understanding}.
\newblock \emph{Preprint}, arXiv:2406.09781.

\bibitem[{Yang et~al.(2024)Yang, Yang, Hui et~al.}]{qwen2}
An~Yang, Baosong Yang, Binyuan Hui, et~al. 2024.
\newblock Qwen2 technical report.
\newblock \emph{arXiv preprint arXiv:2407.10671}.

\bibitem[{Yang et~al.(2023{\natexlab{a}})Yang, Liu, and Wang}]{yang2023fingpt}
Hongyang Yang, Xiao-Yang Liu, and Christina~Dan Wang. 2023{\natexlab{a}}.
\newblock Fingpt: Open-source financial large language models.
\newblock \emph{arXiv preprint arXiv:2306.06031}.

\bibitem[{Yang et~al.(2023{\natexlab{b}})Yang, Yue, and He}]{yang2023autogpt}
Hui Yang, Sifu Yue, and Yunzhong He. 2023{\natexlab{b}}.
\newblock \href {https://arxiv.org/abs/2306.02224} {Auto-gpt for online decision making: Benchmarks and additional opinions}.
\newblock \emph{Preprint}, arXiv:2306.02224.

\bibitem[{Yang et~al.(2020)Yang, Liu, Zhou, Bian, and Liu}]{yang2020qlibaiorientedquantitativeinvestment}
Xiao Yang, Weiqing Liu, Dong Zhou, Jiang Bian, and Tie-Yan Liu. 2020.
\newblock \href {https://arxiv.org/abs/2009.11189} {Qlib: An ai-oriented quantitative investment platform}.
\newblock \emph{Preprint}, arXiv:2009.11189.

\bibitem[{Yu and Yuan(2011)}]{YU2011367}
Jianfeng Yu and Yu~Yuan. 2011.
\newblock \href {https://doi.org/10.1016/j.jfineco.2010.10.011} {Investor sentiment and the mean–variance relation}.
\newblock \emph{Journal of Financial Economics}, 100(2):367--381.

\bibitem[{Yu et~al.(2023)Yu, Li, Chen, Jiang, Li, Zhang, Liu, Suchow, and Khashanah}]{yu2023finmem}
Yangyang Yu, Haohang Li, Zhi Chen, Yuechen Jiang, Yang Li, Denghui Zhang, Rong Liu, Jordan~W. Suchow, and Khaldoun Khashanah. 2023.
\newblock \href {https://arxiv.org/abs/2311.13743} {Finmem: A performance-enhanced llm trading agent with layered memory and character design}.
\newblock \emph{Preprint}, arXiv:2311.13743.

\bibitem[{Zhang et~al.(2024)Zhang, Zhao, Xia, Sun, Sun, Qin, Li, Zhao, Zhao, Cai, Zheng, Wang, and An}]{zhang2024multimodal}
Wentao Zhang, Lingxuan Zhao, Haochong Xia, Shuo Sun, Jiaze Sun, Molei Qin, Xinyi Li, Yuqing Zhao, Yilei Zhao, Xinyu Cai, Longtao Zheng, Xinrun Wang, and Bo~An. 2024.
\newblock \href {https://arxiv.org/abs/2402.18485} {A multimodal foundation agent for financial trading: Tool-augmented, diversified, and generalist}.
\newblock \emph{Preprint}, arXiv:2402.18485.

\end{thebibliography}

\appendix

\section{Overview of Appendix}
We have nearly 15 pages of this appendix, comprising the following subsections for the convenience of readers:

\noindent \textbf{More details about of our framework}
\begin{itemize}[leftmargin=*]
    \item \textbf{\nameref{app:agent}}: This section provides comprehensive instructions on tools, memory, and other details.
    \item \textbf{\nameref{app:prompt}}: This section details the prompt templates used for various tasks within our framework.
    \item \textbf{\nameref{app:Profile}}: A thorough exposition presenting detailed profiles of each agent.

     \item \textbf{\nameref{app:stra}}: This section elaborates on the methodology employed for constructing the strategy pool, which is pivotal for evaluating and refining trading strategies within the QuantAgents framework.
\end{itemize}

\noindent \textbf{More details about of our setting}
\begin{itemize}[leftmargin=*]
    \item \textbf{\nameref{app:prudex}}: An evaluation benchmark assessing performance across multiple dimensions.
    \item \textbf{\nameref{app:dataset}}: Includes details of our datasets.
    \item \textbf{\nameref{app:metrics}}: Includes the calculation of associated metrics.
    \item \textbf{\nameref{app:baseline}}: Comprehensive descriptions of our competitors.
\end{itemize}

\noindent \textbf{More additional experiments}
\begin{itemize}[leftmargin=*]
    \item \textbf{\nameref{app:ablation}}: Additional experiments focusing on ablation study.
    \item \textbf{\nameref{app:aapl}}: This section presents a performance comparison for a single-asset scenario, focusing on Apple Inc. (AAPL) stock from 2021 to 2023, to evaluate the effectiveness of QuantAgents against baseline models.
    \item \textbf{\nameref{app:chinaal}}: This section describes the performance of QuantAgents in real-world trading scenarios within the Chinese market, covering Q1-Q3 of 2024. 
\end{itemize}

\section{Definitions of Single Agent}
\label{app:agent}
In this section, we will provide a comprehensive overview of the composition and execution process of a single agent, designed to simulate the human decision-making process in investments.Each agent comprises a range of financial analysis tools, along with definitions for action, memory, profile, reflection and the execution workflow.

\subsection{Tool}
The tool module $\mathcal{T}$ encompasses a comprehensive suite of technical and analytical tools for investment decision-making, including:

\begin{itemize}[leftmargin=*]
    \item $t_1$: \textbf{Technical Indicator Analysis}, providing analysis of traditional technical indicators such as moving averages, relative strength index, and others;
    \item $t_2$: \textbf{Sentiment Analysis from Social Media}, gauging market sentiment through the analysis of social media platforms;
    \item $t_3$: \textbf{Algorithmic Trading Strategies}, employing algorithms to identify trading opportunities and execute trades;
    \item $t_4$: \textbf{Regulatory Change Impact Analysis}, assessing the potential impact of regulatory changes on the market;
    \item $t_5$: \textbf{Economic Indicator Forecasting}, predicting future economic conditions by analyzing leading economic indicators;
    \item $t_6$: \textbf{Corporate Earnings Analysis}, scrutinizing financial reports to evaluate corporate performance;
    \item $t_7$: \textbf{ NASDAQ-100 Index Component Tracking}, monitoring the performance of individual components within the NASDAQ-100 Index;
    \item $t_8$: \textbf{Sector Performance Evaluation}, assessing the performance of different industry sectors for sector-specific investment decisions;
    \item $t_9$: \textbf{Risk-Adjusted Return Analysis}, measuring the return of an investment in relation to its risk;
    \item $t_{10}$: \textbf{Portfolio Diversification Tools}, aiding in the strategic distribution of investments across various asset classes;
    \item $t_{11}$: \textbf{Central Bank Policy Analysis}, interpreting the implications of central bank policies on currency values and economic conditions;
    \item $t_{12}$: \textbf{Global Macroeconomic Trend Analysis}, examining large-scale economic trends and their impact on global markets;
    \item $t_{13}$: \textbf{Currency Pair Correlation Matrix}, studying the correlation between different currency pairs for informed trading decisions;
    \item $t_{14}$: \textbf{Interest Rate Differential Analysis}, analyzing the effects of interest rate differentials on currency exchange rates;
    \item $t_{15}$: \textbf{Asset Allocation Optimization}, strategically allocating investments to maximize returns and minimize risk;
    \item $t_{16}$: \textbf{Risk Management Frameworks}, employing frameworks to identify, assess, and mitigate investment risks;
    \item $t_{17}$: \textbf{Portfolio Stress Testing}, simulating the impact of extreme market conditions on the portfolio to evaluate its resilience;
    \item $t_{18}$: \textbf{Derivatives Strategy Formulation}, creating strategies involving derivatives to hedge risks and enhance returns;
    \item $t_{19}$: \textbf{Fund Performance Evaluation}, measuring and assessing the performance of investment funds against benchmarks and objectives;
    \item $t_{20}$: \textbf{FinReport}, generating detailed financial reports to provide insights into company and market performance;
    \item $t_{21}$: \textbf{Trend Forecasting}, predicting future market trends based on historical data and predictive analytics;
    \item $t_{22}$: \textbf{Volatility Assessment Tool}, analyzing market volatility to better inform investment decisions;
    \item $t_{23}$: \textbf{Simulation Optimization Toolkit}, optimizing trading strategies through simulation techniques;
    \item $t_{24}$: \textbf{Strategy Analysis Suite}, providing comprehensive analysis of investment strategies for performance evaluation;
    \item $t_{25}$: \textbf{RiskAnalyzer Toolkit}, assessing and quantifying various risk factors in the investment portfolio;
    \item $t_{26}$: \textbf{Risk Score Assessment Tool}, calculating a risk score to guide investment decisions based on the overall risk profile.
\end{itemize}

\subsection{Action}
The specific actions $\mathcal{A}$ that agent can execute include:
\begin{itemize}[leftmargin=*]
    \item $a_1$: \textbf{Buy/Sell/Hold the current assets}, making decisions on whether to acquire new assets, divest existing ones, or maintain the current position;
    \item $a_2$: \textbf{Adjust the quantity and price of securities to be bought or sold}, fine-tuning the volume and pricing strategy for securities transactions;
    \item $a_3$: \textbf{Set or modify trading stop-loss, take-profit, and other trading strategy conditions}, implementing or revising parameters for automated trading strategies to manage risk and lock in profits;
    \item $a_4$: \textbf{Adjust the risk exposure of the investment portfolio}, allocating budget weights and modifying the portfolio to achieve the desired level of risk exposure;
    \item $a_5$: \textbf{Execute Asset Allocation}, strategically distributing investment capital across various asset classes to optimize the portfolio's risk and return profile;
    \item $a_6$: \textbf{Initiate Risk Assessment Protocols}, beginning the process of evaluating potential risks and determining the appropriate measures to mitigate them;
    \item $a_7$: \textbf{Authorize Capital Deployment}, approving the use of funds for investment opportunities in line with the asset allocation strategy;
    \item $a_8$: \textbf{Enforce Compliance with Regulatory Standards}, ensuring that all investment activities adhere to the legal and regulatory framework governing financial markets.
\end{itemize}
\begin{itemize}[leftmargin=*]
    \item $a_1$: \textbf{Buy/Sell/Hold the current assets}, making decisions on whether to acquire new assets, divest existing ones, or maintain the current position;
    \item $a_2$: \textbf{Adjust the quantity and price of securities to be bought or sold}, fine-tuning the volume and pricing strategy for securities transactions;
    \item $a_3$: \textbf{Set or modify trading stop-loss, take-profit, and other trading strategy conditions}, implementing or revising parameters for automated trading strategies to manage risk and lock in profits;
    \item $a_4$: \textbf{Adjust the risk exposure of the investment portfolio}, allocating budget weights and modifying the portfolio to achieve the desired level of risk exposure;
    \item $a_5$: \textbf{Execute Asset Allocation}, strategically distributing investment capital across various asset classes to optimize the portfolio's risk and return profile;
    \item $a_6$: \textbf{Enforce Compliance with Regulatory Standards}, ensuring that all investment activities adhere to the legal and regulatory framework governing financial markets;
    \item $a_7$: \textbf{Rebalance Portfolio}, adjusting the portfolio to maintain the desired asset allocation and risk profile in response to market changes;
    \item $a_8$: \textbf{Conduct Market Scanning}, continuously monitoring the market for new investment opportunities and potential risks;
    \item $a_9$: \textbf{Initiate Hedging Strategies}, implementing hedging techniques such as derivatives to protect the portfolio against adverse market movements;
    \item $a_{10}$: \textbf{Generate Financial Reports}, producing detailed reports on portfolio performance, risk exposure, and compliance status for internal and external stakeholders.
\end{itemize}
\subsection{Memory}
For an agent the Memory module $\mathcal{M}$ is designed to store and manage three main types of information: 
\begin{itemize}[leftmargin=*]
    \item \textbf{Market Information Memory} ($\mathcal{M}{I}$). This memory component stores historical data, including stock prices, financial news, and economic indicators, all of which are relevant to the investment decisions made by the agents.
    \item \textbf{Strategy Memory} ($\mathcal{M}{S}$). This memory component contains a comprehensive analysis of strategies employed in both simulated trading environments and real-world trading scenarios, allowing the agents to refine their approaches based on past performance.
    \item \textbf{Report Memory} ($\mathcal{M}_{R}$). This memory component comprises in-depth analyses of markets, industries, and companies, serving as a reference for the agents when generating investment reports and making informed decisions.
\end{itemize}

\subsection{Profile}
The profile module provides a comprehensive specification of an agent's identity, responsibilities, and operational boundaries, ensuring a well-structured and controlled environment for intelligent investment decision-making. The agent's Profile includes the following main components:
\begin{itemize}[leftmargin=*]
    \item \textbf{Basic Information}: This encompasses the agent type, background information and role assignments.
    \item \textbf{Action Permissions}: The set of actions that the agent is authorized to perform.
    \item \textbf{Tool Permissions}: The collection of analytical tools and resources accessible to the agent for decision-making.
    \item \textbf{Market Information Permissions}: The scope of market data and news that the agent is allowed to access.
    \item \textbf{Team Background}: Provides an overview of the team's structure, the agents' collaborative roles, and the collective goals within the QuantAgents framework.
\end{itemize}

\subsection{Reflection}
The reflection process represents the core decision-making component of the agent. It leverages the LLM foundation to analyze the input information and generate a reflective output based on the specific task objective.
Let $\mathcal{I}_t$ denote the set of inputs available to the agent at time $t$. The reflection process can be formulated as:
\begin{equation}
    \begin{gathered}
        R_t^{\gamma}=\mathcal{R}(\gamma,\mathcal{I}_t)
    \end{gathered}
\end{equation}
Where $\mathcal{R}(\cdot)$ represents reflection function, $\gamma$ represents the current reflective task. The output $R_t^{\gamma}$ is the agent's reflection at time $t$.

The reflection process can be further decomposed into multiple stages, representing the agent's thought process:
i) Information Preprocessing: The agent filters and organizes the input information $\mathcal{I}_t$ based on relevance and importance;
ii)Context Understanding: The agent analyzes the preprocessed information to develop a comprehensive understanding of the current market context;
iii)Goal Alignment: The agent aligns its analysis with the specific investment objectives and constraints defined in its Profile;
iv)Strategy Formulation: Based on the context understanding and goal alignment, the agent formulates potential investment strategies and evaluates their risks and rewards;
v)Decision Making: The agent selects the most suitable strategy, considering its accumulated experience, and generates the final Reflection output $R_t^{\gamma}$.
The Reflection process is iterative and continuously updated as new information becomes available, enabling the agents to adapt their investment decisions dynamically in response to changing market conditions.
\section{Prompt Templates for Various Tasks}\label{app:prompt}
In this section, we detail the prompt templates used for various tasks within the QuantAgents framework. These templates are essential for guiding the Large Language Model (LLM) to generate accurate and context-specific outputs, supporting the decision-making processes of different agents.

\begin{itemize}[leftmargin=*]

    \item \textbf{Market Analysis:} For tasks related to market analysis, the prompt template focuses on delivering a comprehensive evaluation of current market conditions, including key financial indicators and recent news. An example template is:
    
    \begin{center}
    \fcolorbox{black}{gray!10}{\parbox{0.9\linewidth}{
    <Prompt Template> \\
    You are \{Emily Profile\}, a Market Analyst. Today's market overview includes the following data: \{Current Prices\}, \{Recent News\}. Utilize the available financial analysis tools to generate insights. Present the results in JSON format, including key metrics such as \{Tool Results\}.
    }}
    \end{center}

    \item \textbf{Strategy Development:} For strategy development tasks, the prompt template is designed to capture detailed information about strategy formulation and performance evaluation. An example template is:
    
    \begin{center}
    \fcolorbox{black}{gray!10}{\parbox{0.9\linewidth}{
    <Prompt Template> \\
    You are \{Bob Profile\}, a Strategy Analyst. The current strategy is defined by \{Strategy Parameters\}, and the simulation results are as follows: \{Simulation Data\}. Optimize the strategy to enhance \{Performance Metrics\} and provide recommendations in JSON format, highlighting areas for improvement.
    }}
    \end{center}

    \item \textbf{Risk Management:} For risk management tasks, the prompt template emphasizes the assessment and mitigation of financial risks. An example template is:
    
    \begin{center}
    \fcolorbox{black}{gray!10}{\parbox{0.9\linewidth}{
    <Prompt Template> \\
    You are \{Dave Profile\}, a Risk Analyst. Current risk metrics include \{Risk Indicators\}, and recent risk events are \{Risk Events\}. Evaluate the overall risk exposure and propose effective risk mitigation strategies. Output should be in JSON format, with detailed recommendations for risk management.
    }}
    \end{center}

    \item \textbf{Investment Decision:} For investment decision tasks, the prompt template is designed to guide the evaluation of potential investment opportunities based on market conditions and strategic objectives. An example template is:
    
    \begin{center}
    \fcolorbox{black}{gray!10}{\parbox{0.9\linewidth}{
    <Prompt Template> \\
    You are \{Otto Profile\}, an Investment Manager. Evaluate the following investment opportunities: \{Investment Options\}. Considering the current market data \{Market Data\}, determine the optimal investment strategy and provide a JSON-formatted recommendation outlining the suggested actions.
    }}
    \end{center}

\end{itemize}

\section{Profiles of Agents}~\label{app:Profile}
This appendix provides an exhaustive outline of the QuantAgents team member profiles, which are integral to our investment decision-making simulation. The profiles are articulated using XML, chosen for its flexibility and robustness in structuring and representing complex data. XML's self-descriptive nature and the ability to define custom tags make it an ideal choice for encoding the intricate details of each agent's profile. It allows for a high degree of customization and scalability, which is essential for simulating a dynamic and complex environment such as a hedge fund's investment strategy.
\noindent
\begin{figure*}[htbp]
  \centering 
  {\textbf{\large The Profile of Investment Manager Otto}} 
  \rule{\textwidth}{1pt} 
  \small
  \begin{lstlisting}[ breaklines=True, numbers=none]
<profile>
<name>Otto</name>
<description>You are Otto, the Investment Manager who leads the investment team and oversees the entire portfolio. Your responsibility is to make final investment decisions, ensuring that the portfolio is well-diversified and aligned with the fund's objectives. You integrate insights from other agents to craft a cohesive investment strategy that balances risk and return. Your leadership ensures that the portfolio remains adaptive to changing market conditions, and your strategic vision drives the overall success of the investment fund.</description>
<basicInformation>
    <agentType>Portfolio Management Agent</agentType>
    <role>Investment Manager</role>
    <responsibleFor>Overseeing and managing the investment portfolio</responsibleFor>
    <roleAssignment>Final decision-making and strategy oversight.</roleAssignment>
</basicInformation>
<actionPermissions>
    <action>MakeFinalInvestmentDecisions</action>
    <action>AllocateInvestmentBudget</action>
    <action>ApproveStrategies</action>
    <action>MonitorPortfolioPerformance</action>
    <action>AdjustPortfolioAllocation</action>
    <action>EngageInRiskManagement</action>
</actionPermissions>
<toolPermissions>
    <tool>Asset Allocation Optimization</tool>
    <tool>Risk-Adjusted Return Analysis</tool>
    <tool>Central Bank Policy Analysis</tool>
    <tool>Interest Rate Differential Analysis</tool>
    <tool>Derivatives Strategy Formulation</tool>
    <tool>Portfolio Diversification Tools</tool>
</toolPermissions>
<marketInformationPermissions>
    <scope>Historical Trading Data</scope>
    <scope>Real-time Market Data</scope>
    <scope>Portfolio Performance Data</scope>
    <scope>Economic Indicators</scope>
    <scope>Market Sentiment Analysis</scope>
</marketInformationPermissions>
<teamBackground>
    <description>Otto, as the Investment Manager in the QuantAgents framework, coordinates the activities of all agents and ensures the alignment of strategies with the overall investment goals. His role is central to the fund's success, as he integrates diverse insights into a unified strategy and manages the portfolio to maximize returns and manage risks.</description>
</teamBackground>
</profile>
  \end{lstlisting}
  \rule{\textwidth}{1pt} 
\end{figure*}

\hfill
\noindent
\begin{figure*}[htbp]
  \centering 
    \paragraph{The Profile of Strategy Analyst Bob}\mbox{}
    \rule{\textwidth}{1pt}
    \small
    \begin{lstlisting}[language=XML, breaklines=true, numbers=none]
    <profile>
    <name>Bob</name>
    <description>You are Bob, a Strategy Analyst specializing in the development and testing of quantitative trading strategies. Your expertise lies in designing and simulating various investment strategies, using advanced modeling techniques to optimize performance in diverse market conditions. Your decisions are data-driven, leveraging historical and real-time data to refine and validate strategies before they are deployed in live trading. You also collaborate closely with other agents to ensure that strategies align with overall portfolio goals, and you continuously iterate on strategies based on feedback and performance analysis.</description>
    <basicInformation>
        <agentType>Strategy Development Agent</agentType>
        <role>Strategy Analyst</role>
        <responsibleFor>Developing and testing quantitative strategies</responsibleFor>
        <roleAssignment>Strategy design and simulation.</roleAssignment>
    </basicInformation>
    <actionPermissions>
        <action>DevelopStrategy</action>
        <action>SimulateStrategy</action>
        <action>AdjustStrategyParameters</action>
        <action>AnalyzeStrategyPerformance</action>
        <action>OptimizeStrategy</action>
        <action>DeployStrategy</action>
    </actionPermissions>
    <toolPermissions>
        <tool>Technical Indicator Analysis</tool>
        <tool>Strategy Analysis Suite</tool>
        <tool>Simulation Optimization Toolkit</tool>
        <tool>Volatility Assessment Tool</tool>
        <tool>RiskAnalyzer toolkit</tool>
        <tool>Economic Indicator Forecasting</tool>
    </toolPermissions>
    <marketInformationPermissions>
        <scope>Historical Trading Data</scope>
        <scope>Real-time Market Data</scope>
        <scope>Portfolio Performance Data</scope>
    </marketInformationPermissions>
    <teamBackground>
        <description>The QuantAgents framework functions as a sophisticated asset management system, integrating multiple specialized agents to achieve optimal investment outcomes. Bob, as the Strategy Analyst, collaborates with other agents like Otto, the Investment Manager, to align strategy development with overall portfolio management goals, ensuring robust and adaptive trading systems.</description>
    </teamBackground>
    </profile>
    \end{lstlisting}
    \rule{\textwidth}{1pt} 
\end{figure*}
\hfill

\noindent
\begin{figure*}[htbp]
  \centering 
    \paragraph{The Profile of Risk Control Analyst Dave}\mbox{}
    \rule{\textwidth}{1pt}
    \small
    \begin{lstlisting}[ breaklines=true, numbers=none]
    <profile>
    <name>Dave</name>
    <description>You are Dave, a Risk Control Analyst focused on the comprehensive evaluation and mitigation of investment risks. Your role is to monitor the portfolio's risk exposure, assess the risk implications of trading strategies, and implement measures to safeguard against potential losses. Your expertise in risk management tools and frameworks allows you to identify and address vulnerabilities within the investment portfolio, ensuring that the overall risk remains within acceptable limits. You work closely with other agents to ensure that risk considerations are integral to all investment decisions.</description>
    <basicInformation>
        <agentType>Risk Management Agent</agentType>
        <role>Risk Control Analyst</role>
        <responsibleFor>Risk monitoring and mitigation</responsibleFor>
        <roleAssignment>Risk assessment and control.</roleAssignment>
    </basicInformation>
    <actionPermissions>
        <action>EvaluateRiskExposure</action>
        <action>ImplementRiskControls</action>
        <action>MonitorPortfolioRisk</action>
        <action>TriggerRiskAlerts</action>
        <action>AdjustRiskParameters</action>
        <action>PerformStressTesting</action>
    </actionPermissions>
    <toolPermissions>
        <tool>Risk Management Frameworks</tool>
        <tool>Portfolio Stress Testing</tool>
        <tool>RiskAnalyzer toolkit</tool>
        <tool>Risk Score Assessment tool</tool>
        <tool>Regulatory Change Impact Analysis</tool>
        <tool>Economic Indicator Forecasting</tool>
    </toolPermissions>
    <marketInformationPermissions>
        <scope>Historical Trading Data</scope>
        <scope>Real-time Market Data</scope>
        <scope>Portfolio Performance Data</scope>
        <scope>Portfolio Risk Data</scope>
        <scope>Market Volatility Data</scope>
    </marketInformationPermissions>
    <teamBackground>
        <description>In the QuantAgents framework, Dave plays a crucial role in maintaining the balance between risk and return by continuously monitoring and adjusting the portfolio's risk profile. Working in conjunction with other agents, Dave ensures that risk management is a core component of all investment strategies, contributing to the overall stability and success of the portfolio.</description>
    </teamBackground>
    </profile>
    \end{lstlisting}
    \rule{\textwidth}{1pt} 
\end{figure*}
\hfill

\noindent
\begin{figure*}
  \centering 
    \paragraph{The Profile of Market Analyst Emily}\mbox{}
    \rule{\textwidth}{1pt}
    \small
    \begin{lstlisting}[language=XML, breaklines=true, numbers=none]
    <profile>
    <name>Emily</name>
    <description>You are Emily, a Market Analyst dedicated to conducting thorough research on markets, industries, and companies. Your work involves gathering and analyzing vast amounts of data to produce detailed reports that inform investment decisions. Your insights into market trends, sector performance, and company fundamentals are critical for identifying potential investment opportunities and risks. You also provide timely updates and recommendations to ensure that the investment team is well-informed about the latest market developments.</description>
    <basicInformation>
        <agentType>Market Research Agent</agentType>
        <role>Market Analyst</role>
        <responsibleFor>Conducting market research and analysis</responsibleFor>
        <roleAssignment>Research and reporting.</roleAssignment>
    </basicInformation>
    <actionPermissions>
        <action>ConductMarketResearch</action>
        <action>AnalyzeIndustryTrends</action>
        <action>EvaluateCompanyPerformance</action>
        <action>GenerateInvestmentReports</action>
        <action>ProvideMarketUpdates</action>
        <action>RecommendInvestmentActions</action>
    </actionPermissions>
    <toolPermissions>
        <tool>Sector Performance Evaluation</tool>
        <tool>Corporate Earnings Analysis</tool>
        <tool>Trend Forecasting</tool>
        <tool>Fund Performance Evaluation</tool>
        <tool>Global Macroeconomic Trend Analysis</tool>
        <tool>FinReport</tool>
    </toolPermissions>
    <marketInformationPermissions>
        <scope>Market Data</scope>
        <scope>Industry Reports</scope>
        <scope>Company Financial Data</scope>
    </marketInformationPermissions>
    <teamBackground>
        <description>Emily, as the Market Analyst within the QuantAgents framework, is responsible for providing the team with in-depth market intelligence. Her research supports the decision-making process by delivering accurate and actionable insights, making her a key contributor to the success of the overall investment strategy.</description>
    </teamBackground>
    </profile>
    \end{lstlisting}
    \rule{\textwidth}{1pt} 
\end{figure*}
\hfill

\section{Construction of the Strategy Pool}
\label{app:stra}
This section elaborates on the methodology employed for constructing the strategy pool, referenced in the Strategy Development Meeting section. The strategy pool is pivotal for evaluating and refining trading strategies within the QuantAgents framework.

\begin{itemize}[leftmargin=*]

    \item \textbf{Strategy Pool Definition:} The strategy pool encompasses a diverse array of trading strategies derived from permutations of various indices and parameters. This comprehensive array ensures a broad exploration of potential strategies, facilitating the identification of those most effective under varying market conditions.

    \item \textbf{Permutations of Indices and Parameters:} The construction of the strategy pool involves generating permutations across multiple indices and trading parameters. Each permutation represents a unique combination of factors such as technical indicators, risk management rules, and trading signals. This methodology enables a thorough evaluation of strategies across different market environments.

    \item \textbf{Data Utilization:} Historical and real-time market data form the basis for simulating the performance of each strategy within the pool. This data includes stock prices, trading volumes, economic indicators, and other pertinent financial metrics. By applying strategies to this data, we assess their effectiveness and make informed decisions regarding their potential deployment.

    \item \textbf{Performance Evaluation and Selection:} Strategies within the pool are rigorously evaluated based on key performance metrics, including return on investment (ROI), risk-adjusted returns, and maximum drawdown. Strategies demonstrating superior performance metrics are selected for further testing and refinement. This evaluation ensures that only the most promising strategies advance to the subsequent stages of development.

    \item \textbf{Dynamic Updates:} The strategy pool is subject to continuous updates based on new market data and performance feedback. This adaptive approach ensures that the pool remains relevant and responsive to evolving market conditions, integrating new insights and methodologies into strategy development.

\end{itemize}

The strategic construction and maintenance of the strategy pool are crucial for the effective functioning of the QuantAgents framework. 

\section{PRUDEX Evaluation Benchmark}
\label{app:prudex}

\begin{table*}[ht]
\centering
\caption{Summary of evaluation measures for the PRUDEX-Compass framework.}
\label{tab:outer_summary}
\begin{adjustbox}{max width=\textwidth}
\begin{tabular}{@{}lll@{}}
\toprule
\textbf{Axes} & \textbf{Measures} & \textbf{Descriptions} \\
\midrule
Profitability & Profit & Assesses the capability of FinRL methods to accumulate market capital. \\
 & Alpha Decay & Indicates the decline in investment strategy effectiveness over time due to market changes. \\
 & Equity Curve & Represents the portfolio value's progression over time. \\
\midrule
Risk-Control & Risk & Evaluates the level of risk assumed by FinRL methods. \\
 & Risk-adjusted Profit & Normalizes profit against the risk factors, including volatility and downside risk. \\
 & Extreme Market & Measures the performance of FinRL methods during unpredictable market events. \\
\midrule
Universality & Country & Covers performance across a spectrum of financial markets globally. \\
 & Asset Type & Includes a variety of asset classes in the evaluation. \\
 & Time-Scale & Accounts for different trading frequencies in the analysis. \\
\midrule
Diversity & t-SNE & Utilizes statistical methods to visualize the diversity of data points. \\
 & Entropy & Applies information theory to measure the diversity of trading behaviors. \\
 & Correlation & Examines the interrelation between assets to assess strategy diversity. \\
 & Diversity Heatmap & Graphically displays investment decision diversity across assets. \\
\midrule
Reliability & Performance Profile & Provides a visual summary of the empirical score distribution of FinRL methods. \\
 & Variability & Analyzes the consistency of performance across varying conditions. \\
 & Rolling Window & Evaluates the adaptability of FinRL methods over sequential time periods. \\
 & Rank Comparison & Offers a comparative ranking of methods based on performance metrics. \\
\midrule
Explainability & - & This axis is reserved for future discussions on the interpretability of FinRL models. \\
\bottomrule
\end{tabular}
\end{adjustbox}
\end{table*}

PRUDEX-Compass is a systematic evaluation framework for methods in financial markets, consisting of six axes with a total of 17 measures. Each axis represents a critical aspect of the evaluation, and the measures provide detailed insights into the performance of the methods across these aspects.
The following is PRUDEX's specific explanation of the dimensions it assesses.

\textbf{Profitability.} In line with the primary goal of Quantitative Trading (QT) to maximize profits, this criterion assesses the capacity of Financial Reinforcement Learning (FinRL) methods to accumulate market capital. It extends beyond mere returns to encompass the stability and consistency of strategies in achieving substantial profits.

\textbf{Risk-Control.} Given the inherent trade-off between profitability and risk in financial contexts, this element is pivotal. It reflects the industry's emphasis on managing both systemic and unsystematic risks, which is crucial for FinRL method evaluation.

\textbf{Universality.} The financial market's complexity, with its myriad assets and varying temporal and stylistic dimensions, poses a challenge. This criterion evaluates the versatility of FinRL methods in delivering satisfactory performance across diverse trading environments. It aligns with contemporary machine learning trends, such as transfer and meta-learning.

\textbf{Diversity.} Financial diversification, a strategy to spread capital across different assets to mitigate risk, is highlighted by this axis. It addresses the current oversight in FinRL method evaluation concerning diversity, which is essential for bolstering profitability and managing risk.

\textbf{Reliability.} The performance variability of RL methods, their sensitivity to various factors, including random seeds and market fluctuations, can impede reliability—critical for high-stakes financial applications. This axis focuses on the reliability of RL methods within the quantitative trading domain.

\textbf{Explainability.} Trust in a model is paramount for its adoption. Explainability encompasses techniques that clarify a model's behavior, aiding users in understanding model effectiveness under different market conditions and in rectifying erroneous actions. In the regulated financial sector, this is particularly important for model accountability, oversight, and auditing.

In our study, the QuantAgents method has undergone evaluation using the PRUDEX-Compass and has achieved superior scores across all dimensions, showcasing its robustness and effectiveness in financial market simulations.
\section{Details of Dataset Setup}~\label{app:dataset}
This appendix provides a detailed explanation of the technical indicators created in the dataset for this paper, along with their calculation methods and significance. The technical indicators are as follows, a total of 60 indicators.

\textbf{Delta:} Delta is a term from options trading that measures the rate of change of an option's price relative to the underlying asset's price. In the context of stock technical analysis, it might be adapted to reflect sensitivity to market factors.

\textbf{Permutation (Zero-based):} In stock analysis, permutation could be used to analyze sequences of price movements or market conditions, though its application is not standard.

\textbf{Log Return:} Log return is calculated using the natural logarithm and is given by:
\[ \text{Log Return} = \ln\left(\frac{\text{Price at time } t}{\text{Price at time } t-1}\right) \]

\textbf{Max in Range:} This is the highest stock price within a specified range.

\textbf{Min in Range:} This is the lowest stock price within the same range.

\textbf{Middle:} The middle value is the average of the close, high, and low prices:
\[ \text{Middle} = \frac{\text{Close} + \text{High} + \text{Low}}{3} \]

\textbf{Comparison Operators:} These operators are used to establish conditions based on the relationship between data points:
\begin{itemize}[leftmargin=*]
  \item \( \leq \) (less than or equal to)
  \item \( \geq \) (greater than or equal to)
  \item \( < \) (less than)
  \item \( > \) (greater than)
  \item \( = \) (equal to)
  \item \( \neq \) (not equal to)
\end{itemize}

\textbf{Count (Both Backward \( c \) and Forward \( fc \)):} This refers to the tally of occurrences of a condition or event in a dataset, observed both retrospectively and prospectively.

\textbf{Cross:} Crosses indicate when one moving average crosses another, signifying potential trend changes:
\begin{itemize}[leftmargin=*]
  \item \textit{Upward Cross} - bullish signal when a short-term average crosses above a long-term average.
  \item \textit{Downward Cross} - bearish signal when a short-term average crosses below a long-term average.
\end{itemize}

\textbf{SMA (Simple Moving Average):} The SMA is the average of a selected number of time periods and is calculated as follows:
\[ \text{SMA} = \frac{1}{n} \sum_{i=1}^{n} \text{Price}_i \]
where \( n \) is the number of periods and \( \text{Price}_i \) is the price in period \( i \).

\textbf{EMA (Exponential Moving Average):} The EMA gives more weight to recent prices, and it is calculated using the formula:
\[ \text{EMA} = \left( \frac{\text{Price}_t - \text{EMA}_y}{1 + y} \right) + \text{EMA}_y \]
where \( \text{Price}_t \) is the current price, \( \text{EMA}_y \) is the EMA of the previous period, and \( y \) is the number of periods.

\textbf{MSTD (Moving Standard Deviation):} The moving standard deviation measures the average amount by which prices deviate from the SMA over a period and is calculated as:
\[ \text{MSTD} = \sqrt{\frac{1}{n-1} \sum_{i=1}^{n} (\text{Price}_i - \text{SMA})^2} \]

\textbf{MVAR (Moving Variance):} The moving variance is similar to the moving standard deviation but represents the squared deviations:
\[ \text{MVAR} = \frac{1}{n-1} \sum_{i=1}^{n} (\text{Price}_i - \text{SMA})^2 \]

\textbf{RSV (Raw Stochastic Value):} The RSV is used in the calculation of the Stochastic Oscillator and is given by:
\[ \text{RSV} = \frac{\text{Close}_t - \text{Low}_n}{\text{High}_n - \text{Low}_n} \times 100 \]
where \( \text{Close}_t \) is the closing price of the current period, \( \text{Low}_n \) is the lowest price in the last \( n \) periods, and \( \text{High}_n \) is the highest price in the same period.

\textbf{RSI (Relative Strength Index):} The RSI is a momentum oscillator that measures the speed and change of price movements:
\[ \text{RSI} = 100 - \left(100 \div \left(1 + \frac{\text{Average Gains}}{\text{Average Losses}}\right)\right) \]

\textbf{KDJ (Stochastic Oscillator):} The KDJ is a combination of three lines: K, D, and J, which are derived from the RSV and smoothed to identify trends:
\[ \text{K} = \text{SMA}(\text{RSV}, 3) \]
\[ \text{D} = \text{SMA}(\text{K}, 3) \]
\[ \text{J} = 3 \times \text{K} - 2 \times \text{D} \]

\textbf{Bolling (Bollinger Band):} Bollinger Bands consist of a simple moving average plus and minus a standard deviation, providing a measure of volatility:
\[ \text{Upper Band} = \text{SMA} + k \times \text{MSTD} \]
\[ \text{Lower Band} = \text{SMA} - k \times \text{MSTD} \]
where \( k \) is a constant that determines the width of the bands.

\textbf{MACD (Moving Average Convergence Divergence):} The MACD is a trend-following momentum indicator that shows the relationship between two EMAs:
\[ \text{MACD} = \text{EMA}(12) - \text{EMA}(26) \]
with a signal line, usually a 9-day EMA of the MACD line.

\textbf{CR (Energy Index):} The CR is a volume-based indicator that measures the difference between high and low prices to assess market strength:
\[ \text{CR} = \frac{\text{Sum of middle values}}{\text{Sum of high-low ranges}} \]

\textbf{WR (Williams Overbought/Oversold index):} The WR is a momentum indicator that compares the closing price to the high and low range over a period:
\[ \text{WR} = \frac{\text{Highest High} - \text{Close}}{\text{Highest High} - \text{Lowest Low}} \times -100 \]

\textbf{CCI (Commodity Channel Index):} The CCI is designed to measure the difference between an asset's current price and its average price over a certain period:
\[ \text{CCI} = \frac{\text{Price}_t - \text{MA}(\text{Price}_t, n)}{\text{MD}(\text{Price}_t, n) / 0.015} \]
where \( \text{MA} \) is the moving average, \( \text{MD} \) is the mean deviation, and \( n \) is the period.

\textbf{TR (True Range):} The True Range is a measure of market volatility and is defined as the maximum of the following:
\begin{align*}
 \text{TR} = \max(\text{High}_t - \text{Low}_t, |\text{High}_t - \text{Close}_{t-1}|, &\\|\text{Low}_t - \text{Close}_{t-1}|) 
\end{align*}
where \( \text{High}_t \) and \( \text{Low}_t \) are the high and low prices for the current period, and \( \text{Close}_{t-1} \) is the closing price of the previous period.

\textbf{ATR (Average True Range):} The Average True Range is a moving average of the True Range, typically calculated over 14 periods:
\[ \text{ATR} = \text{SMA}(\text{TR}, 14) \]

\textbf{DMA (Difference of Moving Averages):} The DMA is the difference between two moving averages, such as a 10-day and a 50-day:
\[ \text{DMA} = \text{MA}(10) - \text{MA}(50) \]

\textbf{DMI (Directional Moving Index):} The DMI is a set of indicators that measure the strength of a price move in a particular direction.

\textbf{+DI (Positive Directional Indicator):} The +DI indicates the strength of upward moves:
\[ +DI = \frac{\text{H} - \text{L}_m}{\text{TD}} \]

\textbf{-DI (Negative Directional Indicator):} The -DI indicates the strength of downward moves:
\[ -DI = \frac{\text{L}_m - \text{L}}{\text{TD}} \]
where \( \text{H} \) is the high for the period, \( \text{L} \) is the low, \( \text{L}_m \) is the lowest low for the period, and \( \text{TD} \) is the True Range.

\textbf{ADX (Average Directional Movement Index):} The ADX is a measure of the strength of a trend, regardless of direction:
\[ \text{ADX} = \frac{14 \times \text{ADX}_{14} + \text{Current ADX}}{15} \]

\textbf{ADXR (Smoothed Moving Average of ADX):} The ADXR is a smoothed version of the ADX, providing a longer-term view:
\[ \text{ADXR} = \frac{\text{ADX} + \text{Previous ADXR}}{2} \]

\textbf{TRIX (Triple Exponential Moving Average):} The TRIX is a momentum indicator that uses a triple smoothing of the EMA:
\begin{align*}
\text{TRIX} 
&= \frac{\text{Triple EMA}( \text{Price}, n) - \text{Previous TRIX}}
{\text{Previous TRIX}} 
\end{align*}

\textbf{TEMA (Triple Exponential Moving Average):} The TEMA is similar to the TRIX but is designed to be less lagging:
\begin{align*}
\text{TEMA} &= 3 \times \text{EMA}(\text{Price}, n) 
\\&- 3 \times \text{EMA}(\text{EMA}(\text{Price}, n), n) \\
&\quad + \text{EMA}(\text{EMA}(\text{EMA}(\text{Price}, n), n), n)
\end{align*}

\textbf{VR (Volume Variation Index):} The VR is a volume-based indicator that compares the volume of up days to the volume of down days:
\[ \text{VR} = \frac{\text{Up Volume} - \text{Down Volume}}{\text{Up Volume} + \text{Down Volume}} \times 100 \]

\textbf{MFI (Money Flow Index):} The MFI is a volume-weighted indicator that compares money flow to price change:
\[ \text{MFI} = \frac{\text{Money Flow}}{\text{Average Money Flow}} \times 100 \]

\textbf{VWMA (Volume Weighted Moving Average):} The VWMA is an average price that gives more weight to volumes:
\[ \text{VWMA} = \frac{\sum (\text{Price} \times \text{Volume})}{\sum \text{Volume}} \]

\textbf{CHOP (Choppiness Index):} The CHOP index measures market volatility and is calculated using the standard deviation of price changes.

\textbf{KER (Kaufman's Efficiency Ratio):} The KER is a momentum indicator that compares the difference between the high and low of the period to the high of the period:
\[ \text{KER} = \frac{\text{High} - \text{Low}}{\text{High}} \]

\textbf{KAMA (Kaufman's Adaptive Moving Average):} The KAMA is a moving average that adapts to market conditions:
\[ \text{KAMA} = \frac{\text{Previous KAMA} \times (1 + \text{ER})}{1 + \text{ER} + \text{Smoothing Factor}} \]
where \( \text{ER} \) is the efficiency ratio.

\textbf{PPO (Percentage Price Oscillator):} The PPO is a momentum oscillator that compares two moving averages:
\[ \text{PPO} = \left( \frac{\text{EMA}(\text{Short Period}) - \text{EMA}(\text{Long Period})}{\text{EMA}(\text{Long Period})} \right) \times 100 \]
\textbf{StochRSI (Stochastic RSI):} The Stochastic RSI applies the stochastics formula to the RSI indicator, resulting in two lines: \%K and \%D. It is calculated as:
\[ \text{\%K} = \frac{\text{RSI} - \text{Min RSI}}{\text{Max RSI} - \text{Min RSI}} \times 100 \]
\[ \text{\%D} = \text{SMA}(\text{\%K}, 3) \]

\textbf{WT (LazyBear's Wave Trend):} LazyBear's Wave Trend is a trend-following indicator that uses a combination of moving averages and price action to identify the trend direction.

\textbf{Supertrend:} The Supertrend indicator, which consists of an Upper Band and a Lower Band, is a volatility-based band that changes direction when the price touches it:
\[ \text{Upper Band} = \text{Previous Close} + \text{ATR} \times \text{Multiplier} \]
\[ \text{Lower Band} = \text{Previous Close} - \text{ATR} \times \text{Multiplier} \]

\textbf{Aroon: Aroon Oscillator} The Aroon Oscillator is the difference between the Aroon Up and Aroon Down lines:
\[ \text{Aroon Oscillator} = \text{Aroon Up} - \text{Aroon Down} \]

\textbf{Z: Z-Score} The Z-Score is a statistical measurement that describes a score's relationship to the mean of a group of scores:
\[ Z = \frac{(X - \mu)}{\sigma} \]
where \( X \) is the value, \( \mu \) is the mean, and \( \sigma \) is the standard deviation.

\textbf{AO (Awesome Oscillator):} The Awesome Oscillator is a momentum indicator that compares the 5-period and 34-period SMAs:
\[ \text{AO} = \text{SMA}(5) - \text{SMA}(34) \]

\textbf{BOP (Balance of Power):} The Balance of Power is a candlestick-based indicator that compares the opening and closing price to the high and low of the period:
\[ \text{BOP} = \left(\frac{\text{Close} - \text{Open}}{\text{High} - \text{Low}}\right) \times \text{Volume} \]

\textbf{MAD (Mean Absolute Deviation):} The Mean Absolute Deviation is a measure of volatility that averages the absolute deviations from a data set's mean:
\[ \text{MAD} = \frac{1}{n} \sum_{i=1}^{n} |X_i - \mu| \]

\textbf{ROC (Rate of Change):} The Rate of Change is a momentum indicator that measures the percentage change in price over a specific period:
\[ \text{ROC} = \left(\frac{\text{Price}_t - \text{Price}_{t-n}}{\text{Price}_{t-n}}\right) \times 100 \]

\textbf{Coppock: Coppock Curve} The Coppock Curve is a combination of a 14-month rate of change and an 11-month weighted rate of change, smoothed by a 10-month average:
\[ \text{Coppock Curve} = \frac{\text{ROC}(14) + 0.15 \times \text{ROC}(11)}{1 + 0.15} \]

\textbf{Ichimoku: Ichimoku Cloud} The Ichimoku Cloud is a comprehensive charting system that includes five lines:
\[ \text{Tenkan-sen} = \text{SMA}(\text{High}, 9) + \text{SMA}(\text{Low}, 9) \]
\[ \text{Kijun-sen} = \text{SMA}(\text{High}, 26) + \text{SMA}(\text{Low}, 26) \]
\[ \text{Senkou Span A} = \frac{\text{Tenkan-sen} + \text{Kijun-sen}}{2} \]

\begin{align*}
\text{Senkou Span B} &= 
   \frac{\text{SMA}(\text{High}, 52) }{2}
       +&\\ \frac{\text{SMA}(\text{Low}, 52)}{2}
\end{align*}

\textbf{CTI (Correlation Trend Indicator):} The Correlation Trend Indicator measures the correlation between two moving averages to identify trend strength:
\[ \text{CTI} = \frac{\text{SMA}(\text{Price}, n) - \text{SMA}(\text{Price}, m)}{\text{SMA}(\text{Price}, n) + \text{SMA}(\text{Price}, m)} \]

\textbf{LRMA (Linear Regression Moving Average):} The Linear Regression Moving Average is a trend-based indicator that uses the least squares method to fit a line to the data:
\[ \text{LRMA} = b_0 + b_1 \times \text{Time} \]
where \( b_0 \) and \( b_1 \) are the intercept and slope coefficients from the linear regression.

\textbf{ERI (Elder-Ray Index):} The Elder-Ray Index consists of two lines, the \%B (bull power) and \%S (bear power), which are calculated as follows:
\[ \%B = \frac{\text{High} + \text{Low} + \text{Close}}{3} \]
\[ \%S = \frac{\text{High} + \text{Low} - \text{Close}}{3} \]

\textbf{FTR (Gaussian Fisher Transform Price Reversals Indicator):} The Fisher Transform is a statistical tool that converts price data into a Gaussian normal distribution to identify potential reversals:
\[ \text{FTR} = \ln\left(\frac{\text{High} + \text{Low}}{2}\right) \]

\textbf{RVGI (Relative Vigor Index):} The Relative Vigor Index is a momentum indicator that uses the difference between two moving averages to assess market vigor:
\[ \text{RVGI} = \frac{\text{MA}(\text{Price}, n) - \text{MA}(\text{Price}, m)}{\text{MA}(\text{Price}, m)} \times 100 \]

\textbf{Inertia (Inertia Indicator):} The Inertia Indicator is a trend-following indicator that measures the rate of change in price momentum:
\[ \text{Inertia} = \frac{\text{Price}_t - \text{Price}_{t-1}}{\text{Price}_{t-1}} \times 100 \]

\textbf{KST (Know Sure Thing):} The KST is a complex indicator that combines various components, including moving averages and rate of change, to identify trends:
\[ \text{KST} = \text{RS} + \text{ROC} + \text{ST} + \text{M} \]
where \( \text{RS} \) is the raw score, \( \text{ROC} \) is the rate of change, \( \text{ST} \) is the signal, and \( \text{M} \) is the percentage of the smoothed data.

\textbf{PGO (Pretty Good Oscillator):} The Pretty Good Oscillator is a momentum indicator that compares the current price to a moving average:
\[ \text{PGO} = \text{Price}_t - \text{MA}(\text{Price}, n) \]

\textbf{PSL (Psychological Line):} The Psychological Line is a trend-following indicator that uses the percentage of bullish and bearish days:
\[ \text{PSL} = \frac{\text{Number of Bullish Days}}{\text{Total Number of Days}} \times 100 \]

\textbf{PVO (Percentage Volume Oscillator):} The Percentage Volume Oscillator compares the volume of up days to the volume of down days as a percentage:
\[ \text{PVO} = \left(\frac{\text{Up Volume} - \text{Down Volume}}{\text{Up Volume} + \text{Down Volume}}\right) \times 100 \]

\textbf{QQE (Quantitative Qualitative Estimation):} The Quantitative Qualitative Estimation is a method that combines quantitative data with qualitative analysis to provide a comprehensive assessment:
\[ \text{QQE} = \text{Quantitative Score} + \text{Qualitative Score} \]

\section{Details of Evaluation Metrics}~\label{app:metrics}

\textbf{Profit Metrics}:
\begin{itemize}[leftmargin=*]
    \item \textbf{Total Return (TR)}: The percentage change of net value over a time horizon \( h \). It is defined as:
    \[ TR = \left( \frac{n_{t+h} - n_t}{n_t} \right) \times 100\% \]
    
    \item \textbf{Annual Return Rate (ARR)}: The average annual profit of a strategy, calculated as the compound annual growth rate of the net value. It is given by:
    \[ ARR = \left( \left( \frac{n_{t+h}}{n_t} - 1 \right )^{\frac{1}{h}}\right) \times 100\% \]
\end{itemize}

\textbf{Risk-Adjusted Profit Metrics}:
\begin{itemize}[leftmargin=*]
    \item \textbf{Sharpe Ratio (SR)}: The return per unit of deviation. It is calculated as:
    \[ SR = \frac{\mathbb{E}[\mathbf{r}] - r_f}{\sigma[\mathbf{r}]} \]
    
    \item \textbf{Sortino Ratio (SoR)}: A variation that considers only the downside risk. It is given by:
    \[ SoR = \frac{\mathbb{E}[\mathbf{r}] - r_f}{\sigma_{down}[\mathbf{r}]} \]
    
    \item \textbf{Calmar Ratio (CR)}: The annualized return per unit of maximum drawdown. It is defined as:
    \[ CR = \frac{\text{ARR}}{MDD} \]
\end{itemize}

\textbf{Risk Metrics}:
\begin{itemize}[leftmargin=*]
    \item \textbf{Volatility (Vol)}: The standard deviation of the return vector \( \mathbf{r} \), measuring the uncertainty of the return rate. It is calculated as:
    \[ Vol = \sqrt{252} \times \sigma[\mathbf{r}] \]
    
    \item \textbf{Maximum Drawdown (MDD)}: The largest single drop from peak to trough before a new peak is achieved. It is defined as:
    \[ MDD = \max_{0 \leq \tau \leq T} \left[ \max_{0 \leq t \leq \tau} \left( \frac{n_t - n_\tau}{n_t} \right) \right] \]
\end{itemize}

\textbf{Diversity Metrics}:
\begin{itemize}[leftmargin=*]
    \item \textbf{Entropy (ENT)}: A measure of the diversity of bets taken by a strategy, calculated using the Shannon entropy formula. It is given by:
    \[ ENT = -\sum_{i=1}^{N} p_i \log(p_i) \]
    
    \item \textbf{Effect Number of Bets (ENB)}: A measure of the effective number of bets that contribute to the portfolio's performance. It is calculated as:
    \[ ENB = \frac{1}{\sum_{i=1}^{N} (p_i \log(p_i))^2} \]
\end{itemize}

\begin{table*}[]
\caption{Performance comparison of different LLM as the backbone for QuantAgents on 9 evaluation metrics. }
\label{tab:llm}
\small
\centering
\normalsize
\setlength{\tabcolsep}{6pt} 
\renewcommand{\arraystretch}{0.99} 
\resizebox{1\textwidth}{!}{ 
\begin{tabular}{cccccccccc}
\hline
\hline
LLM                & ARR(\%)        & TR(\%)          & SR            & CR             & SoR            & MDD(\%)        & VoL(\%)       & ENT           & ENB           \\\hline
ChatGLM3-6B        & 37.32          & 158.99          & 2.14          & 6.89           & 45.98          & 28.56          & 1.62          & 2.41          & 1.22          \\
Llama-2-13b-chat   & 40.38          & 176.66          & 2.35          & 8.08           & 50.77          & 24.15          & \underline{1.51}    & 2.53          & 1.26          \\
Qwen2-72B-Instruct & 44.13          & 199.41          & 2.23          & 8.59           & 49.22          & 24.52          & 1.77          & 2.66          & 1.33          \\
GPT-4-1106-preview & 53.77          & 263.63          & \underline{2.71}    & 8.76           & \underline{60.11}    & 23.79          & 1.61          & 2.79          & 1.38          \\
Claude 3.5 Sonnet  & \underline{57.95}    & \underline{294.07}    & 2.67          & \underline{10.87}    & 53.74          & \underline{22.33}    & 1.76          & \underline{2.86}    & \underline{1.47}    \\
GPT-4o-2024-05-13  & \textbf{58.68} & \textbf{299.55} & \textbf{3.11} & \textbf{11.38} & \textbf{66.94} & \textbf{16.86} & \textbf{1.43} & \textbf{2.97} & \textbf{1.49} \\
\hline
\hline
\end{tabular}}
\end{table*}
\section{Details of Baselines}~\label{app:baseline}

To comprehensively evaluate the performance of QuantAgents in investment decision-making, we selected a variety of classical and cutting-edge baseline models for comparison. These include three classical rule-based quantitative investment strategies (Classical methods): MV, ZMR, and TSM; three reinforcement learning-based financial agents (RL-based methods): SAC, DeepTrader, and AlphaMix+; and three investment methods based on LLM models (LLM-based methods): FinGPT, FinMem, and FinAgent. A brief introduction to each method is provided below:

\begin{itemize}[leftmargin=*]
    \item \textbf{Classical Methods}
        \begin{itemize}[leftmargin=0cm]
            \item \textbf{Mean-Variance (MV)} is a traditional portfolio optimization strategy that seeks to maximize returns for a given level of risk, or equivalently, minimize risk for a given level of expected returns.
            \item \textbf{Z-score Mean Reversion (ZMR)} assumes that asset prices will revert to their mean over time, using Z-scores to measure the deviation from the mean and identify overbought or oversold conditions.
            \item \textbf{Time Series Momentum (TSM)} is a strategy that exploits momentum in financial markets by investing in assets that have performed well in the past and shorting those that have not.
        \end{itemize}
    \item \textbf{RL-based Methods}
        \begin{itemize}[leftmargin=0cm]
            \item \textbf{Soft Actor-Critic (SAC)} is a state-of-the-art off-policy reinforcement learning algorithm that uses entropy regularization to balance exploration and exploitation in trading strategies.
            \item \textbf{DeepTrader} is a deep reinforcement learning method that optimizes investment policy by embedding macro market conditions to dynamically adjust the proportion between long and short funds, aiming to lower the risk of market fluctuations.
            \item \textbf{AlphaMix+} leverages mixture-of-experts and risk-sensitive approaches to make diversified risk-aware investment decisions, focusing on a comprehensive evaluation framework that includes profitability, risk-control, and other critical axes.
        \end{itemize}
    \item \textbf{LLM-based Methods}
        \begin{itemize}[leftmargin=0cm]
            \item \textbf{FinGPT} is an open-source LLM framework that processes textual and numerical inputs to generate insightful financial decisions, offering advantages over traditional strategies.
            \item \textbf{FinMem} is an advanced LLM agent framework for automated trading, optimized through fine-tuning to enhance performance and returns.
            \item \textbf{FinAgent} is a multimodal foundational agent designed for financial trading tasks, incorporating market intelligence and a dual-level reflection module to adapt to market dynamics and improve decision-making processes.
            \item \textbf{HedgeAgents} is a multi-agent financial trading system leveraging LLMs for robust hedging strategies, featuring specialized analysts and a manager coordinating via conferences to optimize returns and risk management.
        \end{itemize}
\end{itemize}

\section{Experiment of Ablation Study}
\label{app:ablation}
\subsection{Effectiveness of Each Conference}
\begin{figure}[htbp]
  \centering
   \includegraphics[width=0.45\textwidth,keepaspectratio]{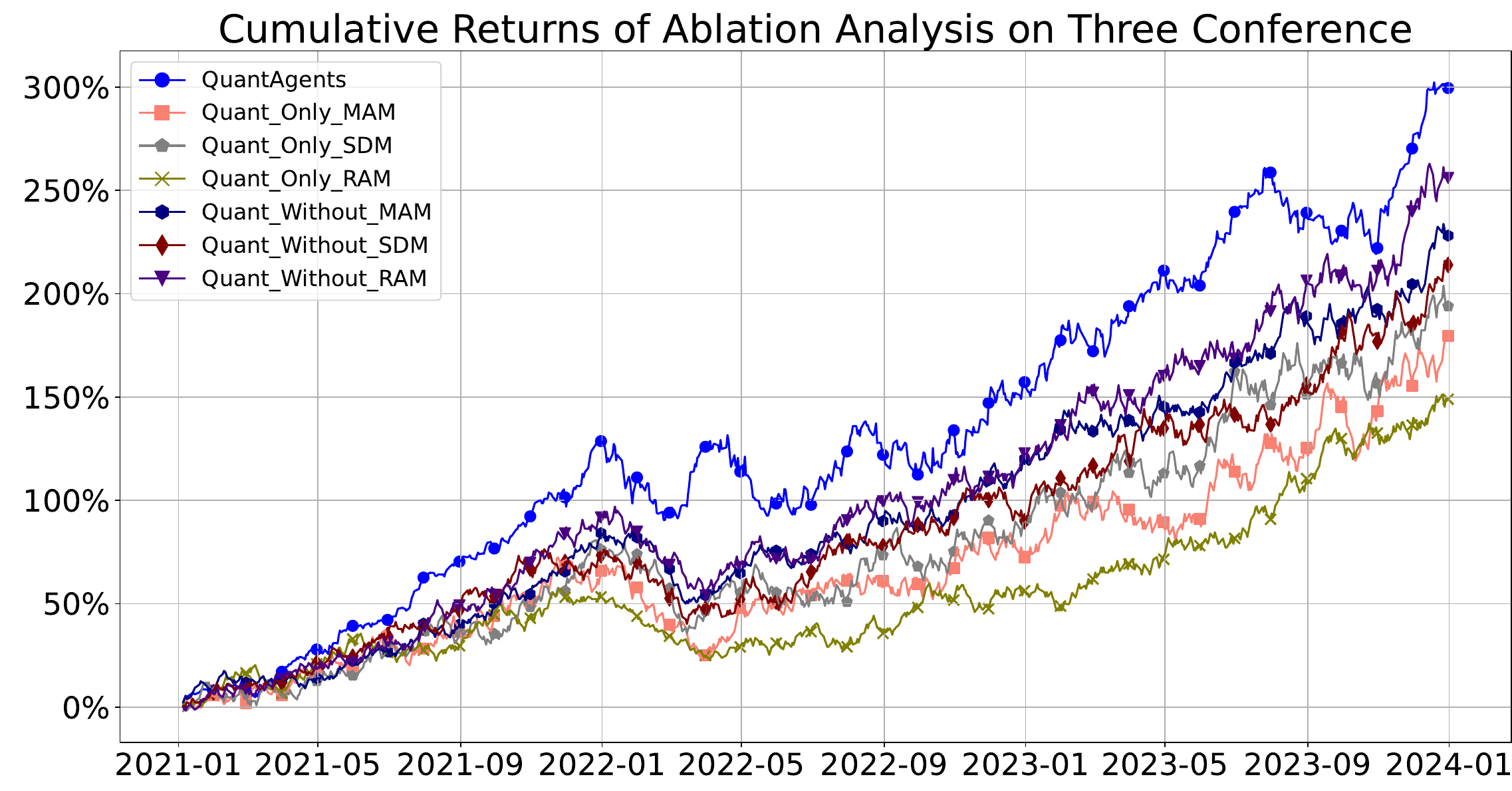}
  \caption{Cumulative Returns of Ablation Analysis on Three Conference}
  \label{fig:CRaba}
\end{figure}

Cumulative returns of ablation analysis on three conference, as shown in Figure \ref{fig:CRaba} .

\subsection{Effectiveness of LLM Backbone}
\begin{figure}[t]
  \centering
   \includegraphics[width=0.47\textwidth,keepaspectratio]{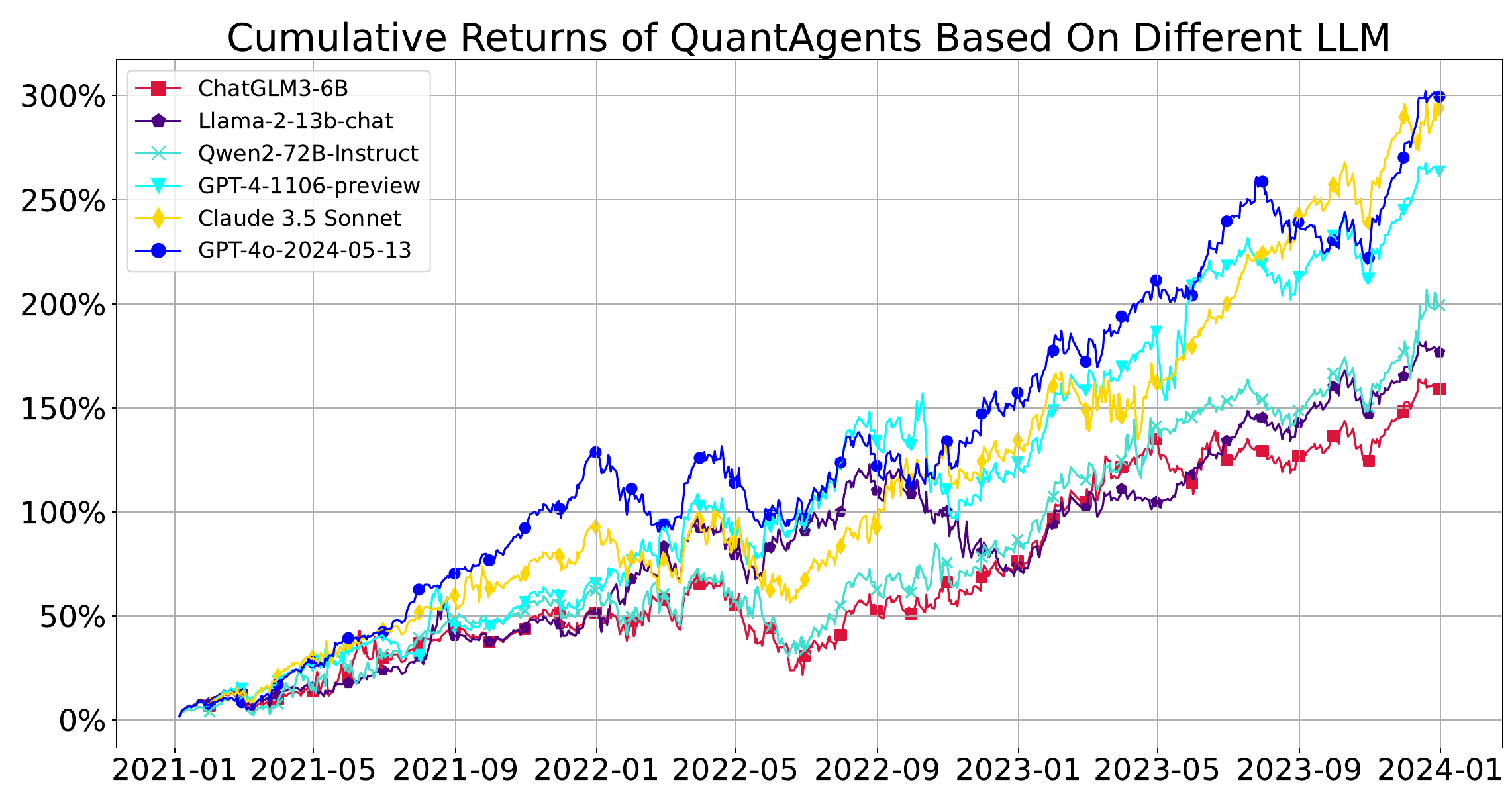}
  \caption{Cumulative Returns of QuantAgents Based on Different LLM}
  \label{fig:CRabb}
\end{figure}

For using different LLM as the backbone for QuantAgents, their experimental results are presented in Table \ref{tab:llm}, and the cumulative returns chart is in Figure \ref{fig:CRabb}.


\section{Single-Asset Performance Comparison}\label{app:aapl}
\begin{figure}[t]
\centering
\includegraphics[width=0.47\textwidth]{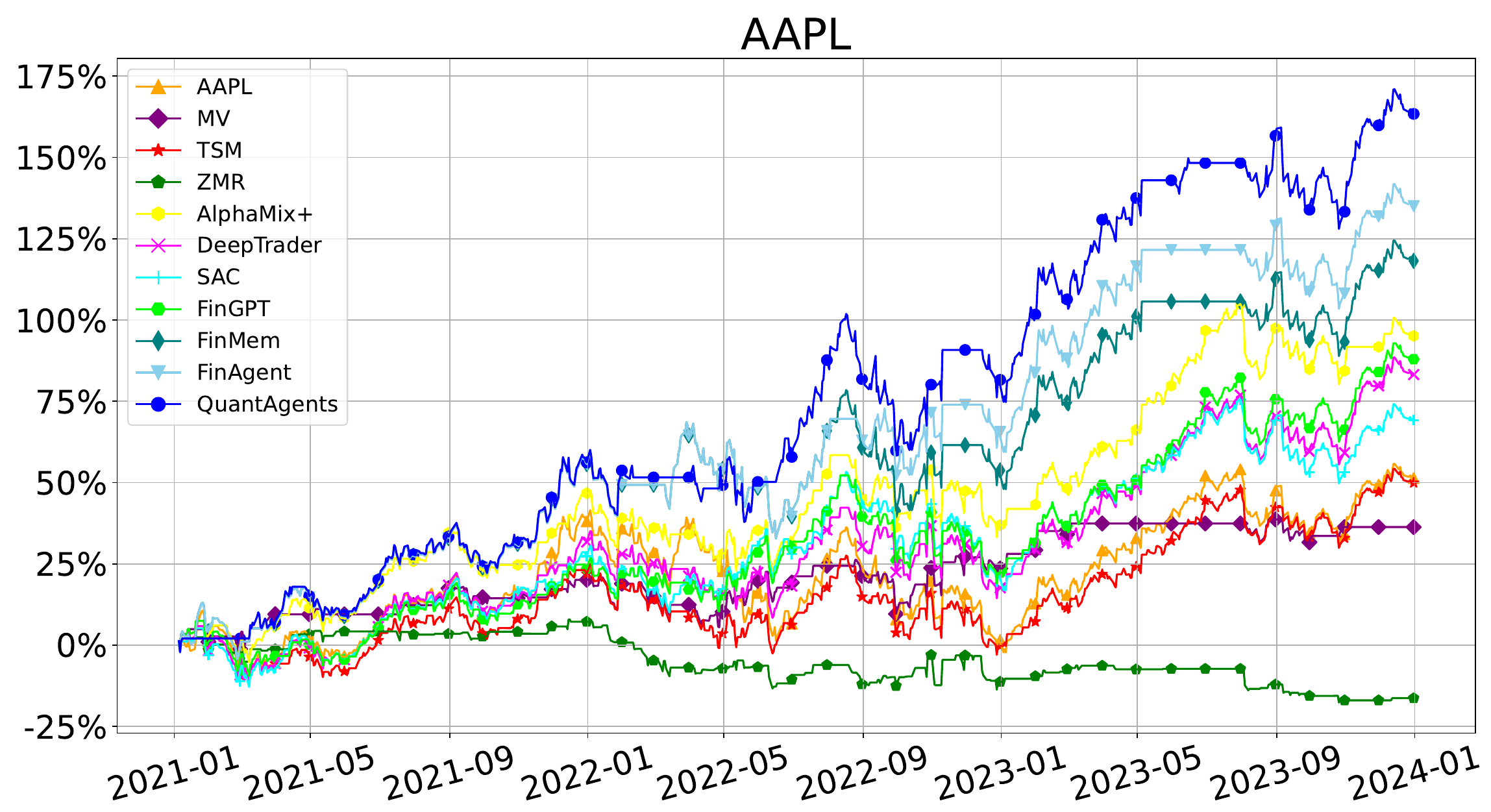}
\caption{Performance comparison of QuantAgents and baseline models on AAPL stock from January 2021 to December 2023.}
\label{fig:aapl_performance}
\end{figure}

To evaluate the effectiveness of all models in a single-asset scenario, we conducted experiments on Apple Inc. (AAPL) stock from 2021-01-01 to 2023-12-31. Figure \ref{fig:aapl_performance} illustrates the performance comparison between QuantAgents and baseline models.

\begin{figure}[t]
    \centering
    \includegraphics[width=0.45\textwidth]
    {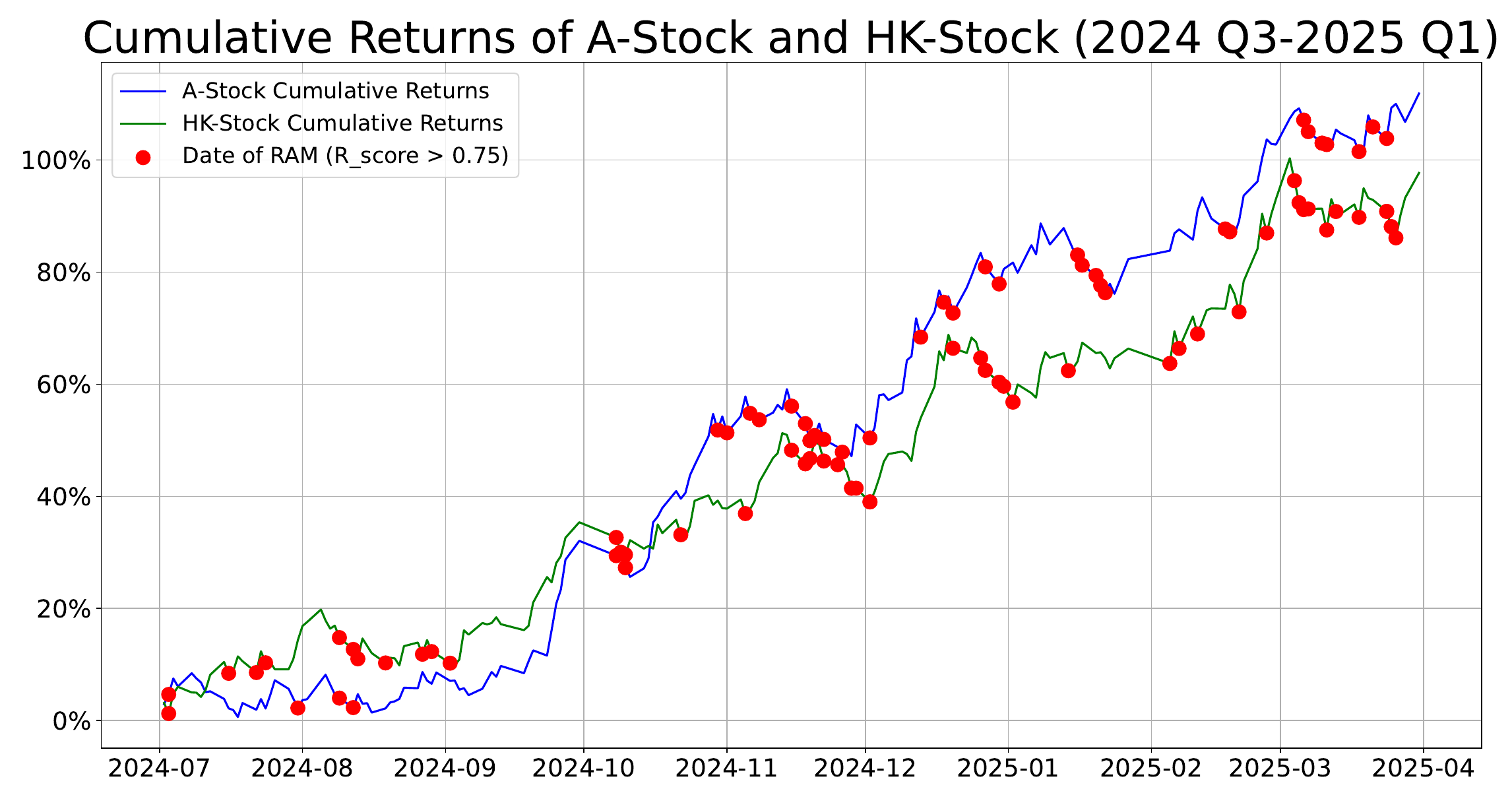}
    \vspace{-0.3cm}
    \caption{Cumulative Returns of QuantAgents during live
trading (24Q3-25Q1). RAM were held 36 times in the
A-stock and 46 times in the HK stock market.}
    \label{fig:live_trading}
    \vspace{-0.5cm}
\end{figure}
The results demonstrate a clear performance hierarchy: 1) RL-based methods outperform rule-based strategies in managing AAPL stock. For instance, SAC achieved a cumulative return of 69.20\% compared to TSM's 49.92\%. This superiority stems from RL methods' ability to adapt to AAPL's high volatility and learn from historical price patterns, enabling more dynamic trading strategies. 2) LLM-based methods surpass RL-based approaches in AAPL trading. For example, FinAgent reached a 135.02\% cumulative return, compared to DeepTrader's 83.22\%. This improvement is attributed to LLMs' capacity to process and interpret AAPL-specific news, earnings reports, and market sentiments, allowing for more informed decision-making in response to company events and sector trends. 3) QuantAgents exhibits superior performance with a 163.38\% cumulative return on AAPL, significantly outperforming all baselines. This exceptional performance stems from its multi-agent architecture, which allows for specialized analysis of AAPL's price movements, market sentiment, and sector trends. 

The integration of advanced LLMs enables QuantAgents to process AAPL-related news and financial reports more effectively. Additionally, the dual reward mechanism enhances QuantAgents' ability to balance risk and return specifically for AAPL stock, resulting in more stable performance during both bullish and bearish periods in the stock.

\section{Empirical Evaluation of QuantAgents in Live Trading}\label{app:chinaal}
\begin{table}[h]
    \centering
    \small
    \setlength{\tabcolsep}{3pt}
    \caption{Performance Metrics of QuantAgents in Live Trading (Q3 2024--Q1 2025).}
    \resizebox{0.45\textwidth}{!}{
    \begin{tabular}{lccc}
        \hline
        \textbf{Market} & \textbf{Total Return (\%)} & \textbf{Sharpe Ratio} & \textbf{Win Rate (\%)} \\
        \hline
        A-stocks       & 111.87                    & 2.02                 & 61.23                 \\
        HK-stocks      & 97.69                     & 1.76                 & 59.71                 \\
        \hline
    \end{tabular}}
    \label{tab:market_performance}
\end{table}

To rigorously validate the efficacy of QuantAgents, we conducted an extensive evaluation of its live trading performance in the A-stock (Shanghai and Shenzhen) and HK-stock (Hong Kong) markets over the period from Q3 2024 to Q1 2025. These markets were selected due to their distinct characteristics: A-stocks exhibit high volatility and liquidity driven by domestic retail investors, while HK-stocks are influenced by international capital flows and stricter regulatory frameworks. This diversity tests QuantAgents' adaptability to varying market dynamics.

The experimental setup involved deploying QuantAgents in a live trading environment with a diversified portfolio, adhering to real-world constraints such as transaction costs and market impact. Risk Alert Meetings (RAM) were convened to monitor and mitigate potential downturns, occurring 36 times for A-stocks and 46 times for HK-stocks, reflecting the latter's higher volatility. Figure~\ref{fig:live_trading} illustrates the cumulative returns over the evaluation period.

QuantAgents achieved superior returns of 111.87\% in the A-stock market, with a Sharpe Ratio of 2.02 and a Win Rate of 61.23\%, demonstrating robust profitability under volatile conditions. In the HK-stock market, it recorded returns of 97.69\%, with a Sharpe Ratio of 1.76 and a Win Rate of 59.71\%, showcasing consistent performance despite international market complexities. These results, detailed in Table~\ref{tab:market_performance}, highlight QuantAgents' exceptional profitability and risk management capabilities across diverse market conditions, underscoring its potential for real-world financial applications.

\section{Conclusions of Appendix}
In this nearly 15 page appendix, we provide additional details about our framework (Section \nameref{app:agent}, \nameref{app:prompt},\nameref{app:Profile}), experimental settings (Section \nameref{app:prudex}, \nameref{app:dataset}, \nameref{app:metrics}, \nameref{app:baseline}, \nameref{app:stra}), and a more additional experiments (Section \nameref{app:ablation}, \nameref{app:aapl},\nameref{app:chinaal}). We hope that our efforts will serve as a source of inspiration for more readers!

\end{document}